%% file: Main.tex
\documentclass[10pt,twocolumn,letterpaper]{article}

\usepackage{iccv}
\usepackage{times}
\usepackage{epsfig}
\usepackage{graphicx}
\usepackage{amsmath}
\usepackage{amssymb}

\usepackage{caption}
\usepackage{subcaption}
\usepackage{graphicx}
\usepackage{amsmath}
\usepackage{amssymb}
\usepackage{booktabs}

\usepackage{CJKutf8}
\usepackage{makecell}
\usepackage{multirow}
\DeclareMathOperator*{\argmax}{arg\,max}

\usepackage{xr-hyper}
\usepackage{comment}
\usepackage{amsmath}
\usepackage{algorithm}
\usepackage{algpseudocode}
\usepackage{listings}

\usepackage[breaklinks=true,bookmarks=false]{hyperref}

\iccvfinalcopy 

\def\iccvPaperID{X}

\ificcvfinal\fi

\begin{document}

\title{Improving Continuous Sign Language Recognition with Cross-Lingual Signs}

\author{Fangyun Wei\\
Microsoft Research Asia\\
{\tt\small fawe@microsoft.com}
\and
Yutong Chen\\
Microsoft Research Asia\\
{\tt\small chenytjudy@gmail.com} \\
}

\maketitle
\ificcvfinal\thispagestyle{empty}\fi

\input{sections/abstract}
\input{sections/introduction}

\input{sections/related_works}

\input{sections/method}
\input{sections/experiments}
\input{sections/conclusion}

{\small
\bibliographystyle{ieee_fullname}
\bibliography{egbib}
}

\appendix
\input{sections/appendix}

\end{document}

%% file: sections/abstract.tex
\begin{abstract}
This work dedicates to continuous sign language recognition (CSLR), which is a weakly supervised task dealing with the recognition of continuous signs from videos, without any prior knowledge about the temporal boundaries between consecutive signs. Data scarcity heavily impedes the progress of CSLR. Existing approaches typically train CSLR models on a monolingual corpus, which is orders of magnitude smaller than that of speech recognition. In this work, we explore the feasibility of utilizing multilingual sign language corpora to facilitate monolingual CSLR. Our work is built upon the observation of cross-lingual signs, which originate from different sign languages but have similar visual signals (e.g., hand shape and motion). The underlying idea of our approach is to identify the cross-lingual signs in one sign language and properly leverage them as auxiliary training data to improve the recognition capability of another. To achieve the goal, we first build two sign language dictionaries containing isolated signs that appear in two datasets. Then we identify the sign-to-sign mappings between two sign languages via a well-optimized isolated sign language recognition model. At last, we train a CSLR model on the combination of the target data with original labels and the auxiliary data with mapped labels. Experimentally, our approach achieves state-of-the-art performance on two widely-used CSLR datasets: Phoenix-2014 and Phoenix-2014T.
\end{abstract}

%% file: sections/introduction.tex
\vspace{-3mm}
\section{Introduction}
\label{sec:intro}
Sign languages are visual-spatial signals for communication among deaf communities. These languages are primarily expressed through hand shape but are also greatly aided by the movement of the body, head, mouth, and eyes. Sign language recognition, which aims to establish communication between hearing people and deaf people, can be roughly categorized into two sorts: isolated sign language recognition (ISLR)~\cite{zuo2023natural,li2020word,MSASL,li2020transferring} and continuous sign language recognition (CSLR)~\cite{koller2017re-sign,camgoz2020sign,koller2019weak,zuo2022c2slr,haozhou2020STMC,twostream_slr}. ISLR is a supervised classification task—it requires models to recognize and classify isolated signs from videos. In contrast, CSLR is a weakly supervised task dedicated to the recognition of continuous signs from videos, without any prior knowledge about the temporal boundaries between consecutive signs. The objective of this work is to develop a CSLR framework, with the assistance of an ISLR model and multilingual corpus.

\begin{figure}
     \centering
    \begin{subfigure}[b]{0.48\textwidth}
         \centering
        \includegraphics[width=\textwidth]{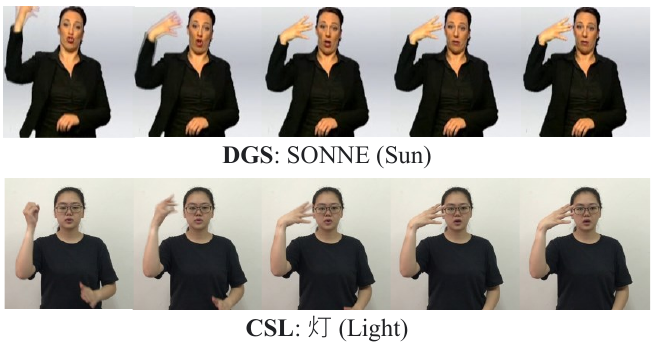}
         \vspace{-7mm}
         \caption{Cross-lingual signs usually have distinct meanings.}
         \label{fig:teaser_B}
     \end{subfigure}
     \hfill
     \vspace{-3mm}
     \begin{subfigure}[b]{0.48\textwidth}
         \centering
         \includegraphics[width=\textwidth]{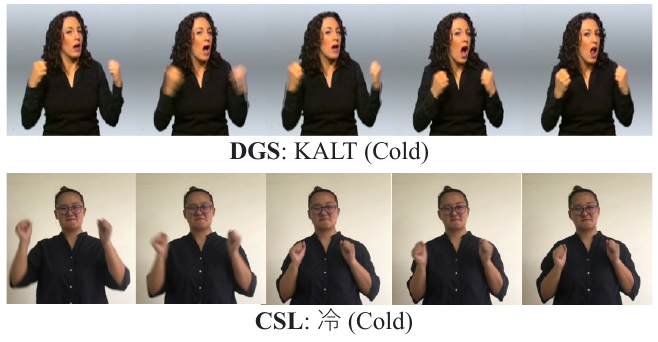}
         \vspace{-7mm}
         \caption{Cross-lingual signs occasionally convey the same meaning.}
         \label{fig:teaser_A}
     \end{subfigure}
     \hfill
    \vspace{-6mm}
    \caption{Cross-lingual signs are those that originate from different sign languages but have similar visual signals (\textit{e.g.} hand shape and motion). We show two examples identified by our approach from a German sign language (DGS) dataset and a Chinese sign language (CSL) dataset.}
    \vspace{-5mm}
    \label{fig:teaser}
\end{figure}

\begin{figure*}[t]
\centering
\includegraphics[width=1.0\linewidth]{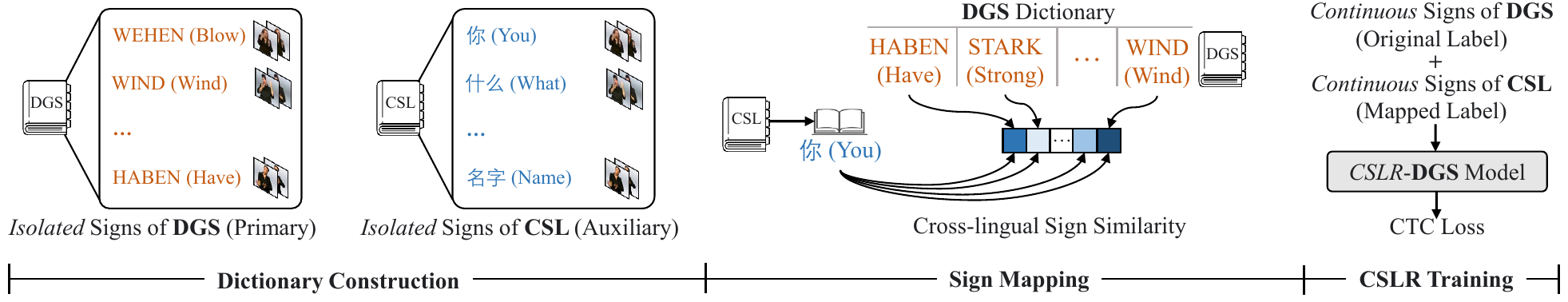}
\caption{Illustration of our approach. The objective is to train a continuous sign language recognition (CSLR) model in DGS with the assistance of a CSL dataset. We first build two sign dictionaries containing isolated signs in DGS and CSL respectively. Then we train an isolated sign language recognition (ISLR) model to identify the CSL-to-DGS mapping for each isolated sign in the CSL dictionary. Finally, the CSLR-DGS model is trained on a combination of the DGS dataset with original labels and the CSL dataset with \emph{mapped} labels through the CTC loss~\cite{graves2006connectionist}.}
\vspace{-4mm}
\label{fig:teaser_C}
\end{figure*}

In contrast to the promising achievements in automatic speech recognition~\cite{li2022recent,amodei2016deep,park2019specaugment,watanabe2018espnet}, the lack of large-scale training data heavily impedes the progress of CSLR. In general, training a satisfying speech recognition model requires thousands of hours of training data~\cite{librilight,panayotov2015librispeech}. However, existing sign language datasets~\cite{P2014,camgoz2018neural,zhou2021improving} are orders of magnitude smaller, containing only fewer than 20 hours of paralleled samples. Previous approaches~\cite{koller2017re-sign,camgoz2020sign,koller2019weak,zuo2022c2slr,haozhou2020STMC,twostream_slr} typically train CSLR models on monolingual corpora such as German sign language (Deutsche Gebärdensprache, DGS) datasets (\textit{e.g.} Phoenix-2014~\cite{P2014} and Phoenix-2014T~\cite{camgoz2018neural}), and Chinese sign language (CSL) datasets (\textit{e.g.} CSL-Daily~\cite{zhou2021improving}). Unfortunately, the limited training data immensely restricts the recognition capacity. One possible technique to alleviate the data scarcity issue is semi-supervised learning, which requires a large volume of unlabeled data in addition to a collection of labeled data. However, existing CSLR datasets are collected in specific domains, \textit{e.g.}, Phoenix-2014~\cite{P2014} and Phoenix-2014T~\cite{camgoz2018neural} are concentrated in the domain of weather forecast. Collecting domain-relevant data for CSLR becomes challenging and unpractical, impeding semi-supervised learning.

Nevertheless, since sign languages are visual languages, it is possible for them to employ the same sign to express either the same or different meanings. If the hypothesis is true, it will be feasible to utilize multilingual sign language corpora to enrich training data. Fortunately, we conduct experiments to confirm the existence of these signs, which are referred to as cross-lingual signs in our work. Cross-lingual signs are those that originate from different sign languages but have similar visual signals (\textit{e.g.}, hand shape and motion). Since most sign languages are mutually unintelligible, cross-lingual signs typically have distinct meanings in different sign languages (Figure~\ref{fig:teaser_B}). Interestingly, we find that they might convey the same meanings occasionally (Figure~\ref{fig:teaser_A}). The discovery of cross-lingual signs inspires us to identify these signs in one sign language and properly leverage them as auxiliary training data to improve the recognition performance of the other.

Consider a scenario where we are going to train a CSLR model in DGS given a primary DGS dataset and an auxiliary dataset in another sign language, \textit{e.g.}, CSL. In general, the size of the auxiliary dataset should exceed that of the primary one. The underlying idea behind our approach is to find the CSL-to-DGS mapping for each isolated sign in the CSL dataset. Figure~\ref{fig:teaser_C} illustrates our solution, which contains three steps: (1) build two sign language dictionaries containing isolated signs that appear in CSL and DGS datasets, respectively; (2) identify the CSL-to-DGS mapping for each isolated sign in CSL dictionary according to the cross-lingual sign similarity calculated by a well-optimized isolated sign language recognition model; (3) train a CSLR model in DGS on the combination of DGS data with original labels and CSL data with mapped labels through the well-established CTC loss~\cite{graves2006connectionist}. It is worth mentioning that it is non-trivial to directly build a sign language dictionary from a CSLR dataset due to the absence of sign boundary annotations. To tackle this problem, we adopt a pre-trained CSLR model to split 
the continuous signs into the isolated ones for dictionary construction.

The contributions of this work can be summarized as:
\vspace{-2mm}
\begin{itemize}
    \item We are the first to utilize a multilingual sign language corpus to facilitate monolingual CSLR based on the finding of cross-lingual signs.
    \vspace{-2mm}
    \item We present a comprehensive solution for seamlessly incorporating an auxiliary dataset—though in another sign language—into the training of CSLR.
    \vspace{-2mm}
    \item Our approach achieves state-of-the-art performance on two widely-used CSLR datasets: Phoenix-2014~\cite{P2014} and Phoenix-2014T~\cite{camgoz2018neural}.
\end{itemize}

%% file: sections/related_works.tex
\section{Related Works}
\noindent\textbf{Sign Language Recognition} can be categorized into isolated sign language recognition (ISLR) and continuous sign language recognition (CSLR). ISLR aims to classify an isolated sign video into a single sign class~\cite{zuo2023natural,li2020word,MSASL,li2020transferring,Albanie2021bobsl}. CSLR aims to word-by-word transcribe a co-articulated sign video into a sign sequence~\cite{koller2017re-sign,camgoz2020sign,twostream_slr,haozhou2020STMC,Cui2017Recurrent,hu2022temporal}. Similar to autonomous speech recognition (ASR)~\cite{yu2016automatic,rabiner1993fundamentals,li2022recent}, CSLR is a weakly supervised sequence-to-sequence task without temporal boundary annotation available and hence is typically trained using CTC loss~\cite{graves2006connectionist}. To address the scarcity of parallel CSLR annotation, Pu et al. estimate the temporal boundaries and stitch video clips to generate pseudo parallel data for data augmentation~\cite{CrossAug_MM2020}. Some methods iteratively optimizes the model with a sequence-level CTC loss and an auxiliary frame-level classification loss~\cite{Pu2019Iterative,Cui2017Recurrent}. Other works exploit multi-stream networks with multi-modal inputs or labels~\cite{STMC_MM,twostream_slr,koller2019weak,Camg2020Multichannel}, or recognize the meanings that signers express through sign language retrieval~\cite{cheng2023cico}. Our CSLR method alleviates data scarcity by leveraging multilingual training data and integrating ISLR models into the CSLR framework to find cross-lingual signs.

\vspace{2mm}
\noindent\textbf{Cross-lingual Transfer} is well explored in speech-to-text recognition~\cite{toshniwal2018multilingual,li2019bytes,kannan2019large,ardila2019common} and spoken language understanding~\cite{johnson2017google,liu2020multilingual, conneau2019unsupervised}. This line of research commonly finds that training a model on multilingual corpora can yield superior performance on low-resource languages compared to training it on a monolingual corpus. In the context of sign language understanding, there are also a few publications investigating cross-lingual transfer~\cite{Gokul2022_MultiSL,Tornay2020_MultilingualSLR,Yin2022_MLSLT}. For sign language translation, MLSLT~\cite{Yin2022_MLSLT} proposes a multilingual translation network with language-specific parameters that outperforms the monolingual baseline. For ISLR, Gokul et al. combine labels from different sign languages based on their word meanings~\cite{Gokul2022_MultiSL}. Despite bringing improvements, this approach overlooks the fact that the same words can be expressed differently across different sign languages. For CSLR, Tornay et al. train a hand movement model using a different sign language resource before optimizing the classifier using the target sign language data in their KL-HMM framework~\cite{Tornay2020_MultilingualSLR}. However, their cross-lingual model falls short of the monolingual reference. Our CSLR method utilizes both target and auxiliary sign language data via a cross-lingual sign mapping and shows improvements over the monolingual baseline.

%% file: sections/method.tex
\section{Method}
\begin{figure*}
     \centering
     \begin{subfigure}[b]{0.95\textwidth}
         \centering
         \includegraphics[width=\textwidth]{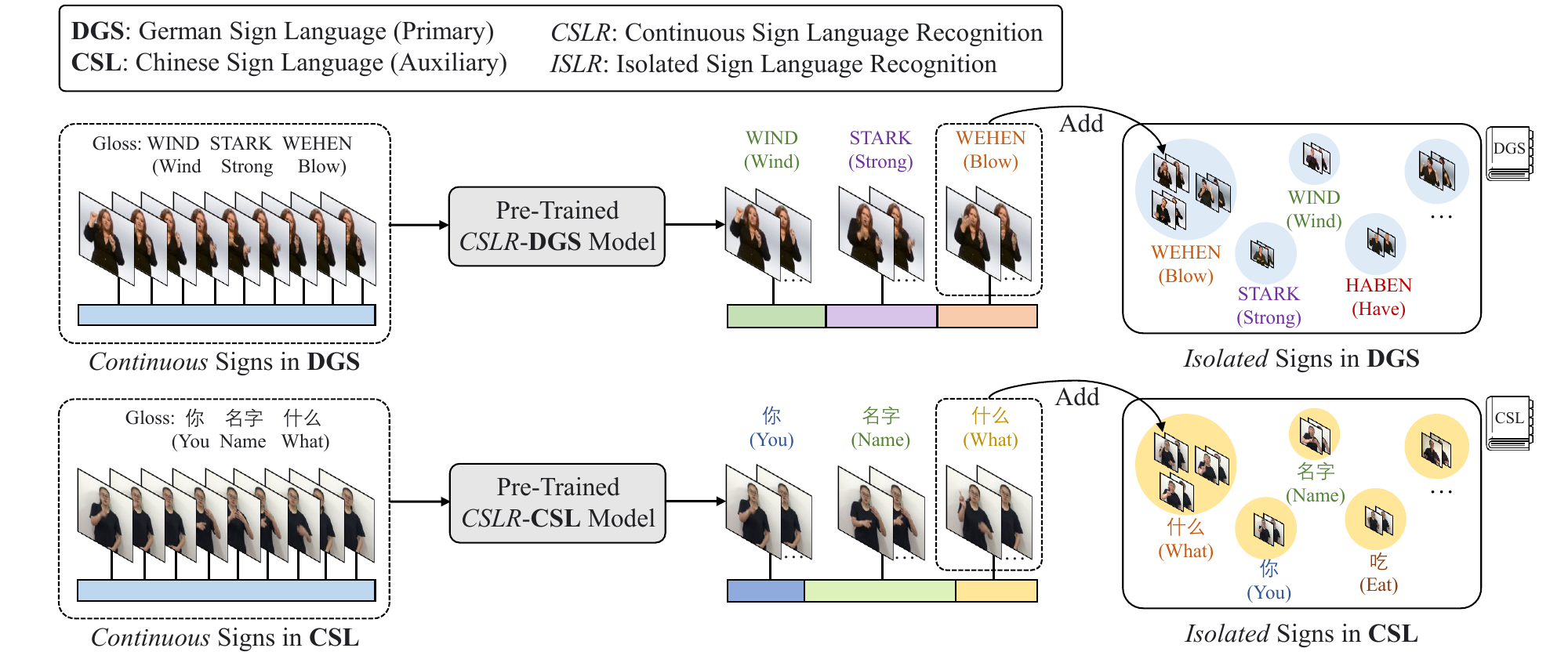}
         \vspace{-6mm}
         \caption{Dictionary construction for DGS and CSL datasets.}
         \label{fig:overview_A}
     \end{subfigure}
     \hfill
     \vspace{3mm}
     \begin{subfigure}[b]{0.95\textwidth}
         \centering
         \includegraphics[width=\textwidth]{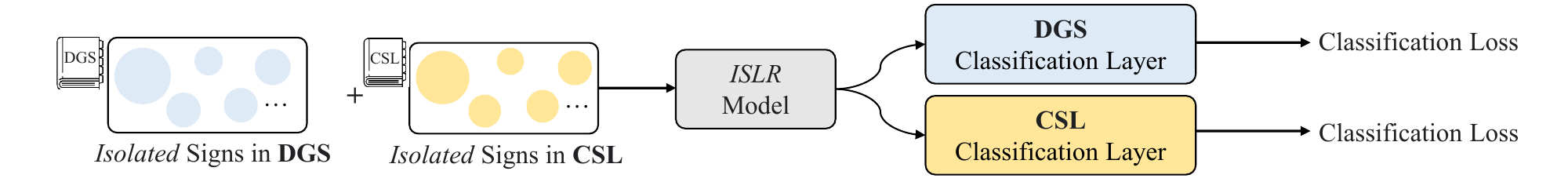}
         \caption{Train a multilingual ISLR model.}
         \label{fig:overview_B}
     \end{subfigure}
     \hfill
      \begin{subfigure}[b]{0.95\textwidth}
         \centering
         \includegraphics[width=\textwidth]{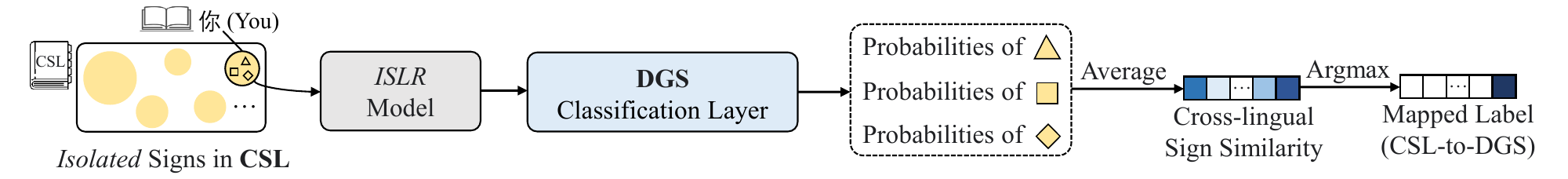}
         \vspace{-5mm}
         \caption{Identify CSL-to-DGS sign mappings through cross-lingual similarities.}
         \label{fig:overview_C}
     \end{subfigure}
     \hfill
     \vspace{3mm}
    \begin{subfigure}[b]{0.95\textwidth}
         \centering
         \includegraphics[width=\textwidth]{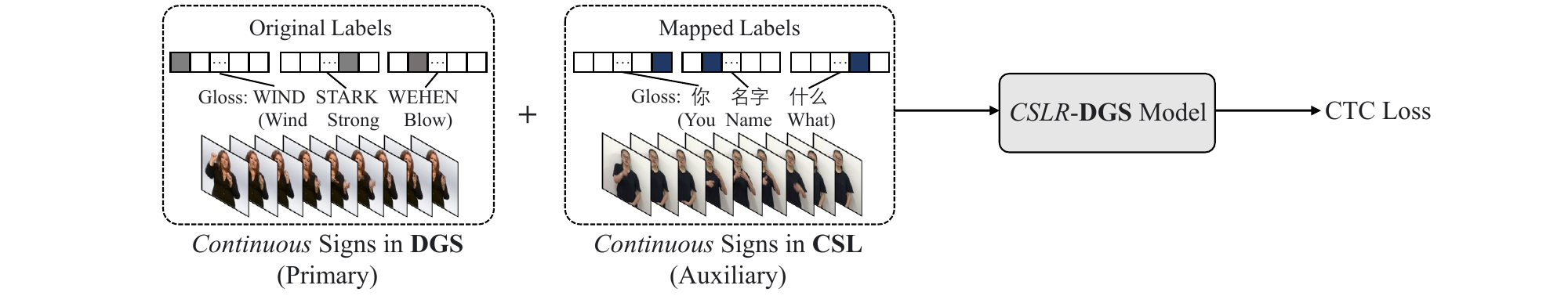}
         \caption{Train a CSLR model in DGS on the combination of the DGS dataset with raw labels and the CSL dataset with mapped labels.}
         \label{fig:overview_D}
     \end{subfigure}
     \hfill
     \vspace{-2mm}
    \caption{Overview of our method. Given a primary dataset in DGS and an auxiliary dataset in CSL, we train a CSLR model in DGS with the assistance of an auxiliary CSL dataset. Our method consists of four steps as illustrated in (a)-(d).}
    \label{fig:overview}
    \vspace{-2mm}
\end{figure*}
\noindent\textbf{Objective.} The goal of continuous sign language recognition (CSLR) is to recognize a sequence of signs from videos. CSLR is a weakly supervised task since the temporal boundaries between consecutive signs are unprovided. In this work, we attempt to utilize multilingual sign language corpus to facilitate monolingual CSLR based on the finding of cross-lingual signs. Concretely, given a \textit{primary} CSLR dataset $\mathcal{D}_{P}$ in \textit{target} sign language, and an \textit{auxiliary} CSLR dataset $\mathcal{D}_{A}$ in \textit{source} sign language, the objective of this work is to train a CSLR model using both $\mathcal{D}_{P}$ and $\mathcal{D}_{A}$ to improve the recognition performance on $\mathcal{D}_{P}$. In general, the size of $\mathcal{D}_{A}$ should exceed that of $\mathcal{D}_{P}$.

\vspace{1mm}
\noindent\textbf{Overview.} Figure~\ref{fig:overview} visualizes the training process of our approach, which consists of three steps: 1) build two sign dictionaries $\mathcal{C}_{P}$ and $\mathcal{C}_{A}$ containing isolated signs which appear in $\mathcal{D}_{P}$ and $\mathcal{D}_{A}$, respectively (Section~\ref{sec:dic_const}); 2) identify source-to-target mapping for each isolated sign in $\mathcal{C}_{A}$ (Section~\ref{sec:sign_map}); 3) train a CSLR model in target sign language using $\mathcal{D}_{P}$ with original labels and $\mathcal{D}_{A}$ with mapped labels (Section~\ref{sec:cslr}).

\subsection{Dictionary Construction}
\label{sec:dic_const}
A sign dictionary $\mathcal{C}$ is a set of isolated signs, each of which has a number of video instances associated with it. Existing CSLR datasets do not provide the corresponding dictionaries. To circumvent this issue, we propose to construct sign dictionaries on CSLR datasets. However, it is not feasible to simply split continuous signs into isolated ones without effort, as sign boundaries are not readily available in CSLR datasets. Fortunately, a well-optimized CSLR model is able to predict the boundaries of signs in addition to its main capability of recognizing a sequence of signs, which inspires us to employ a pre-trained CSLR model to automatically split the continuous signs into isolated ones to construct dictionaries, as shown in Figure~\ref{fig:overview_A}.

Concretely, given a CSLR dataset $\mathcal{D}$ ($\mathcal{D}_P$ or $\mathcal{D}_A$), we first train a CSLR model on it. This model can be any CSLR model~\cite{STMC_MM,zuo2022c2slr,MMTLB_2022,CrossAug_MM2020,twostream_slr} which is able to produce frame-wise predictions. In this work, we adopt the TwoStream-SLR model~\cite{twostream_slr} due to its superior performance. Once the CSLR model is well optimized, given a training video $\boldsymbol{v} \in \mathcal{D}$ containing $T$ frames and its associated ground-truth sequence label $\boldsymbol{s}=(s_1,\ldots, s_N)$ containing $N$ consecutive signs, we use the pre-trained CSLR model to compute the most probable alignment path ${\boldsymbol{\pi}}^*=(\pi_1,\ldots,\pi_T),\pi_t \in \{s_i\}_{i=1}^{N}$ where $\pi_t$ predicts which sign is being performed at the $t$-th frame. 
 
The computation of ${\boldsymbol{\pi}}^*$ can be efficiently implemented via the dynamic time warping (DTW) algorithm~\cite{berndt1994using_dtw}. The details are described in the supplementary material. 
With $\boldsymbol{\pi}^*$ indicating frame-wise prediction, we can split $\boldsymbol{v}$ into $N$ non-overlapped clips where each clip is associated with an isolated sign appeared in $\boldsymbol{s}$. Then we add each clip to the collection of its associated sign in dictionary $\mathcal{C}$. We repeat the above process for each video in $\mathcal{D}$. Finally, we obtain a sign dictionary $\mathcal{C}$ with an alphabet $\mathcal{S}$. The alphabet $\mathcal{S}$ is the set of all signs appearing in  $\mathcal{D}$. In the constructed dictionary $\mathcal{C}$, each sign $s\in \mathcal{S}$ has a collection of isolated sign videos.

We apply two CSLR models trained on $\mathcal{D}_{P}$ and $\mathcal{D}_{A}$ to construct their corresponding sign dictionaries $\mathcal{C}_{P}$ with alphabet $\mathcal{S}_P$, and $\mathcal{C}_{A}$ with alphabet $\mathcal{S}_A$, respectively.

\subsection{Cross-lingual Sign Mapping}
\label{sec:sign_map}
Now we introduce the process of identifying source-to-target mapping for each isolated sign in $\mathcal{S}_{A}$. To measure sign-to-sign similarities, we adopt a simple yet effective approach by training a multilingual isolated sign classifier using the two dictionaries $\mathcal{C}_{A}$ and $\mathcal{C}_{P}$. The unified classifier encodes signs from both dictionaries into a shared embedding space where visually similar signs (cross-lingual signs) are closer to each other than dissimilar ones. By doing so, we can employ the classifier to find visually similar signs across the two sign languages and establish a cross-lingual sign mapping.

\vspace{-4mm}
\subsubsection{Multilingual ISLR}
\vspace{-3mm}
Isolated sign language recognition (ISLR)~\cite{MSASL,li2020word,li2020transferring} aims to classify isolated signs. Unlike prior work which trains ISLR models on monolingual datasets, we train a unified classifier on the combination of $\mathcal{C}_{P}$ and $\mathcal{C}_{A}$ for multilingual prediction. Figure~\ref{fig:overview_B} shows our ISLR model, which consists of a shared vision encoder and two language-aware classification layers. The vision encoder encodes input videos from $\mathcal{C}_{P}$ and $\mathcal{C}_{A}$ into a shared embedding space. The two separate classification layers then project embeddings into two probability distributions over $\mathcal{S}_P$ and $\mathcal{S}_A$ respectively. We train the multilingual ISLR model by minimizing the sum of two cross-entropy losses on $\mathcal{C}_{P}$ and $\mathcal{C}_{A}$. 
Besides, as both isolated training sets are heavily long-tailed, we filter out signs with low frequency to alleviate the class imbalance issue. We find that once optimized, the shared vision encoder can automatically align the embedding spaces of $\mathcal{C}_{P}$ and $\mathcal{C}_{A}$—it draws close sign videos that are visually similar regardless of the data source.

\vspace{-3mm}
\subsubsection{Cross-lingual Sign Mapping}
\label{sec:xlingual_sim}
\vspace{-2mm}
By using the trained multilingual ISLR model which encodes isolated sign videos in a multilingual embedding space, we are now able to identify the cross-lingual sign mappings between the signs in $\mathcal{S}_{A}$ and those in $\mathcal{S}_{P}$. Formally, given a sign $s\in \mathcal{S}_A$ belonging to the auxiliary language, we intend to find its mapped sign $\mathcal{M}_{A\rightarrow P}(s) \in \mathcal{S}_P$ belonging to the primary sign language. We propose two strategies for identifying such mappings, utilizing our multilingual ISLR model, as described below.

\vspace{1mm}
\noindent\textbf{Cross-lingual Prediction.} This is our default mapping strategy as shown in Figure~\ref{fig:overview_C}. Recall that our ISLR model is composed of a shared vision encoder to extract vision features, and two classification layers to classify sign videos from $\mathcal{C}_A$ and $\mathcal{C}_P$ respectively. For the purpose of computing cross-lingual similarities, given a sign video from $\mathcal{C}_A$, we feed it through the vision encoder and then the classification layer of $\mathcal{C}_P$ to predict its probability distribution over $\mathcal{S}_P$. In other words, we can generate a cross-lingual probability for each video in $\mathcal{C}_A$ by switching the classification layer to that of $\mathcal{C}_P$. We denote the cross-lingual probability of $\boldsymbol{v}\in \mathcal{C}_A$ as $p_{A\rightarrow P}(\boldsymbol{v})\in \mathbb{R}^{|\mathcal{S}_P|}$, which predicts the distribution of video $\boldsymbol{v}$ over the alphabet $\mathcal{S}_P$. 
The highest activation in $p_{A\rightarrow P}(\boldsymbol{v})$ indicates the most similar sign in $\mathcal{S}_P$ to the video $\boldsymbol{v}$. To map a sign $s\in \mathcal{S}_A$ to $\mathcal{S}_P$, we average the cross-lingual predictions of its all associated video instances: 
\begin{equation}
    \label{eq:x_sim_cross}
    \begin{split}
\mathcal{M}_{A\rightarrow P}^{\mathrm{cls}}(s) &= \mathrm{argmax}(p_{A\rightarrow P}(s)), \\
   p_{A\rightarrow P}(s) &= \frac{1}{|\mathcal{C}_A(s)|}\sum_{\boldsymbol{v}\in \mathcal{C}_A(s)}p_{A\rightarrow P}(\boldsymbol{v}),~ s \in \mathcal{S}_A,
   \end{split}
\end{equation}
where $\mathcal{C}_A(s)$ denotes the set of videos in $\mathcal{C}_A$ which are associated with the sign $s\in \mathcal{S}_A$. This type of mapping is referred to as a \emph{class-level mapping}. Note that all video instances belonging to the same sign share identical mapping. In addition, we also explore a variant, \emph{instance-level mapping}, which identifies mapping for each individual video instance. We use $\mathcal{M}_{A\rightarrow P}^{\mathrm{ins}} (\boldsymbol{v})$ to denotes the instance-level mapping of a video $\boldsymbol{v} \in \mathcal{C}_A$, which is formulated as:
\begin{equation}
    \label{eq:x_sim_ins}
   \mathcal{M}_{A\rightarrow P}^{\mathrm{ins}}(\boldsymbol{v}) = \mathrm{argmax}(p_{A\rightarrow P}(\boldsymbol{v})),~ \boldsymbol{v} \in \mathcal{C}_A.
\end{equation}

\vspace{1mm}
\noindent\textbf{Dot-product of Weight Matrices.} 
Another way to compute cross-lingual similarity is to utilize the weights of two well-optimized classification layers, which we denote as $\boldsymbol{W}_A \in \mathbb{R}^{|\mathcal{S}_A| \times d}$ and $\boldsymbol{W}_P \in \mathbb{R}^{|\mathcal{S}_P| \times d}$, respectively. The $i$-th row of the weight matrix $\boldsymbol{W}\in \mathbb{R}^{|B| \times d}$ can be interpreted as the learned prototype of the $i$-th sign. Therefore, we can calculate pair-wise similarities between $\mathcal{S}_A$ and $\mathcal{S}_P$ by dot-producting their weight matrices as $\boldsymbol{W}_{A}\boldsymbol{W}_{P}^T \in \mathbb{R}^{|\mathcal{S}_A|\times|\mathcal{S}_P|}$, where the $i$-th row denotes the similarities between the $i$-th sign in $\mathcal{S}_A$ and all signs in $\mathcal{S}_P$. Then we normalize the row vectors via a softmax operation to obtain the mapping from each auxiliary sign to the primary alphabet, which is formulated as:
\begin{equation}
    \label{eq:weight_mat}
    \begin{split}
    \mathcal{M}_{A\rightarrow P}^{\rm{weight}}(s_i) &= \mathrm{argmax}(p_{A\rightarrow P}(s_i)), \\
    p_{A\rightarrow P}(s_i) &= \mathrm{Softmax}((\boldsymbol{W}_{A}\boldsymbol{W}_{P}^T)[i,:]),~ s_i \in \mathcal{S}_A.
    \end{split}
\end{equation}

\vspace{-1mm}
\noindent\textbf{Cross-lingual Mapping.}
Now for each sign $s \in \mathcal{S}_A$, we can compute its mapped sign $\mathcal{M}_{A\rightarrow P}(s) \in \mathcal{S}_P$ by either Eq.~\ref{eq:x_sim_cross} or Eq.~\ref{eq:weight_mat}. In this case, all video instances belonging to the same sign share identical mapping. For the variant defined by Eq.~\ref{eq:x_sim_ins}, we simply map each sign video $\boldsymbol{v} \in \mathcal{C}_A$ at the instance level. This means that video instances belonging to the same sign from the source dataset can be assigned to distinct signs in the target dataset. We visualize some examples in Figure~\ref{fig:teaser} and the supplementary materials.

\subsection{Training CSLR on Multilingual Corpus}
\label{sec:cslr}
Now we could map the labels of sign videos in the auxiliary CSLR dataset $\mathcal{D}_A$ from the source sign language to the target sign language. Specifically, for each continuous sign video $\boldsymbol{v}\in\mathcal{D}_A$ with raw labels $\boldsymbol{s}=(s_1,\ldots, s_N)$, we replace each $s_i$ with its mapped label as described in Section~\ref{sec:sign_map}. We use $\mathcal{D}_{A\rightarrow P}$ to denote the yielded dataset. Then a CSLR model is trained on the combination of $\mathcal{D}_P$ and $\mathcal{D}_{A\rightarrow P}$ as shown in Figure~\ref{fig:overview_D}. Following previous CSLR methods~\cite{twostream_slr,zuo2022c2slr,STMC_MM}, we use CTC loss~\cite{graves2006connectionist} to train a TwoStreamSLR network~\cite{twostream_slr}. Given a video $\boldsymbol{v}$ and its corresponding sequence label $\boldsymbol{s}$, the CTC loss is formulated as:

\begin{equation}
    \label{eq:ctc}
    \mathcal{L}(\boldsymbol{v},\boldsymbol{s}) = -\ln\sum_{\boldsymbol{\pi}\in \Pi (\boldsymbol{s})}{p\left(\boldsymbol{\pi} \middle | \boldsymbol{v}\right)}.
\end{equation}
where $\boldsymbol{\pi}$ is an alignment path and $\Pi(\boldsymbol{s})$ is the set of all possible alignments for $\boldsymbol{s}$. The conditional probability $p\left(\boldsymbol{\pi} \middle | \boldsymbol{v}\right)$ is a product of cross-entropy losses over all time steps. 

Our CSLR model is trained by minimizing
\begin{equation}
    \label{eq:m_ctc}
    \mathcal{L}_{\textrm{CSLR}} = \mathbb{E}_{(\boldsymbol{v},\boldsymbol{s})\sim [\mathcal{D}_P \cup \mathcal{D}_{A\rightarrow P}]_{1:\alpha}} \mathcal{L}(\boldsymbol{v},\boldsymbol{s}),
\end{equation}
where $[]_{1:\alpha}$ denotes that the training samples in a mini-batch are randomly sampled from $\mathcal{D}_P$ and $\mathcal{D}_{A\rightarrow P}$ with a ratio of $1:\alpha$. We set $\alpha<$ 1 to ensure the primary dataset $\mathcal{D}_P$ dominates the training.

%% file: sections/experiments.tex
\section{Experiments}

\subsection{Implementation Details}
\vspace{0mm}
\noindent\textbf{Datasets.} Our experiments involve two German sign language (DGS) datasets namely Phoenix-2014~\cite{P2014} and Phoenix-2014T~\cite{camgoz2018neural}, and a Chinese sign language (CSL) dataset named CSL-Daily~\cite{zhou2021improving}. The Phoenix-2014 dataset is split into train/dev/test with 5672/540/629 videos respectively and contains a total of 1231 signs. The Phoenix-2014T dataset has a split of 7096/519/642 videos and contains 1085 signs. The two DGS datasets share 958 signs in common. The CSL-Daily dataset has a split of 18401/1077/1176 videos and includes 2000 unique signs. 

\vspace{0.5mm}
\noindent\textbf{Settings.} Since the two DGS datasets are smaller in scale and vocabulary size than the CSL dataset, we choose Phoenix-2014/Phoenix-2014T as the primary dataset $\mathcal{D}_{P}$ and CSL-Daily as the auxiliary dataset $\mathcal{D}_{A}$ in our experiments. This ensures that $\mathcal{D}_{A}$ provides sufficient cross-lingual signs to complement the training source of $\mathcal{D}_{P}$. Additionally, we can also use one of the two DGS datasets as the auxiliary dataset for the other. Due to the significant overlap in their vocabularies, we can directly merge their vocabularies without using the cross-lingual mappings described in Section~\ref{sec:sign_map}. In summary, we verify our methodology under four settings (auxiliary dataset$\rightarrow$primary dataset): 1) CSL-Daily$\rightarrow$Phoenix-2014T; 2) CSL-Daily$\rightarrow$Phoenix-2014; 3) CSL-Daily+Phoenix-2014$\rightarrow$Phoenix-2014T; 4) CSL-Daily+Phoenix-2014T$\rightarrow$Phoenix-2014.

\vspace{0.5mm}
\noindent\textbf{Model Architecture.}
Our CSLR network follows the architecture of TwoStream-SLR~\cite{twostream_slr}, which adopts a dual encoder to model RGB videos and keypoint sequences. We conduct ablation studies on SingleStream-SLR with only RGB inputs due to its computational efficiency. More details about the architecture can be found in~\cite{twostream_slr}. We also train separate monolingual TwoStream-SLRs to segment the co-articulated videos for dictionary construction  (Section~\ref{sec:dic_const}). Our ISLR model adopts the same network architecture except that an average pooling layer and a classification layer are appended on top of the network.

\vspace{0.5mm}
\noindent\textbf{Training.} For the ISLR model, we train it for $100$ epochs with a batch size of 32 and a learning rate of 1e-4. During training, we filter out the tail classes with a frequency threshold of $8$ for Phoenix-2014T/Phoenix-2014 and $20$ for CSL-Daily. During inference, we forward all samples to compute their cross-lingual predictions (Section~\ref{sec:xlingual_sim}). For the CSLR model, we follow the training scheme of TwoStream-SLR~\cite{twostream_slr}—training it for $40$ epochs with a batch size of 8 and a learning rate of 1e-3. We set $\alpha$ as 0.2 in Eq~\ref{eq:m_ctc}. We show more details in the supplementary.

\noindent\textbf{Evaluation.} We evaluate our CSLR model on the primary dev/test set using CTC decoding~\cite{graves2006connectionist} with beam width set to $5$. Following prior works~\cite{koller2017re-sign,STMC_MM,camgoz2020sign,twostream_slr}, we use Word Error Rate (WER) as the evaluation metric, which is a normalized edit distance between the prediction and the reference~\cite{graves2006connectionist}. Lower WER indicates higher recognition performance. We run each experiment three times with different random seeds and report the score of the best checkpoint.
\vspace{-2mm}
\subsection{Comparison with State-of-the-art Methods}
\vspace{-1mm}
\noindent\textbf{Phoenix-2014T.} We evaluate our method on the Phoenix-2014 benchmark and compare with previous work in Table~\ref{tab:sota_phoenix2014T}. TwoStream-SLR~\cite{twostream_slr} achieves the leading performance among previous works by using a dual encoder to model both RGB and keypoint inputs. We implement our cross-lingual method using their TwoStream architecture and utilize CSL-Daily as an auxiliary dataset to improve the performance of Phoenix-2014T (CSL-Daily$\rightarrow$Phoenix-2014T). 
We show that incorporating CSL-Daily improves the performance of its monolingual counterpart (TwoStream-SLR) by 0.4/0.7 WER on the dev/test sets. Further, we employ Phoenix-2014 as another auxiliary dataset besides CSL-Daily (CSL-Daily+Phoenix-2014$\rightarrow$Phoenix-2014T). Adding Phoenix-2014 can further improve the performance of Phoenix-2014T, achieving a state-of-the-art with WER of $16.9/18.5$ on the dev/test sets.
\begin{table}[t]
\setlength\tabcolsep{8pt} 
    \centering
    \resizebox{0.99\linewidth}{!}{ 
    \begin{tabular}{l cc}
    \toprule
        Method&Dev&Test\\
        \midrule
        CNN-LSTM-HMMs~\cite{koller2019weak}&22.1&24.1  \\
        SFL~\cite{niu2020stochastic}&25.1&26.1\\
        FCN~\cite{fcn_eccv2020}&23.3&25.1\\
        Joint-SLRT~\cite{camgoz2020sign}&24.6&24.5\\
        SignBT~\cite{zhou2021improving}&22.7&23.9  \\
        MMTLB~\cite{MMTLB_2022}&21.9&22.5 \\
        SMKD~\cite{Hao_2021_ICCV} &20.8&22.4\\
        STMC-R~\cite{STMC_MM}&19.6&21.0\\
        C$^2$SLR~\cite{zuo2022c2slr} &20.2 & 20.4  \\
        TwoStream-SLR ~\cite{twostream_slr}&\underline{17.7}&\underline{19.3}\\
        \midrule
        CSL-Daily$\rightarrow$Phoenix-2014T&17.3&18.6 \\       
        CSL-Daily+Phoenix-2014$\rightarrow$Phoenix-2014T&\textbf{16.9}&\textbf{18.5} \\ 
    \bottomrule
    \end{tabular}}
    \vspace{-2mm}
    \caption{Comparison with previous work on \textbf{Phoenix-2014T} with WER as the evaluation metric. We underline the best results in previous work and bold the best results achieved by our methods.}
    \label{tab:sota_phoenix2014T}
    \vspace{-5mm}
\end{table}

\vspace{-2mm}
\noindent\textbf{Phoenix-2014.} Similar to the experimental setting on Phoenix-2014T, we evaluate our method on the Phoenix-2014 benchmark and compare its performance with previous work in Table~\ref{tab:sota_phoenix2014}. The results on Phoenix-2014 are in line with those on Phoenix-2014T. Leveraging CSL-Daily as an auxiliary (CSL-Daily$\rightarrow$Phoenix-2014) outperforms monolingual TwoStream-SLR~\cite{twostream_slr} by 0.4/0.3 WER. Adding both Phoenix-2014T and CSL-Daily (CSL-Daily+Phoenix-2014T$\rightarrow$Phoenix-2014) reduces the WER to 15.7/16.7.

\begin{table}[t]
\setlength\tabcolsep{8pt} 
    \centering
    \resizebox{0.99\linewidth}{!}{ 
    \begin{tabular}{l cc}
    \toprule
        Method&Dev&Test\\
        \midrule
        SubUNets~\cite{subunet_iccv2017}&40.8&40.7 \\
        IAN~\cite{Pu2019Iterative}&37.1&36.7\\
        ReSign~\cite{koller2017re-sign}&27.1&26.8\\
        CNN-LSTM-HMMs~\cite{koller2019weak}&26.0&26.0  \\
        SFL~\cite{niu2020stochastic}&24.9&25.3\\
        DNF~\cite{dnf_cui} &23.8&24.4\\
        FCN~\cite{fcn_eccv2020}&23.7&23.9\\
        DNF~\cite{dnf_cui} &23.1&22.9\\
        VAC~\cite{Min_2021_ICCV}&21.2&22.3  \\
        LCSA~\cite{zuo22_interspeech} &21.4&21.9\\
        CMA~\cite{CrossAug_MM2020} &21.3&21.9\\
        SMKD~\cite{Hao_2021_ICCV} &20.8&21.0\\
        STMC-R~\cite{STMC_MM}&21.1&20.7\\
        C$^2$SLR~\cite{zuo2022c2slr} & 20.5 & 20.4\\
        TwoStream-SLR~\cite{twostream_slr}&\underline{18.4}&\underline{18.8}\\
        \midrule
       CSL-Daily$\rightarrow$Phoenix-2014 &18.0&18.5\\
       CSL-Daily+Phoenix-2014T$\rightarrow$Phoenix-2014 &\textbf{15.7}&\textbf{16.7} \\      
    \bottomrule
    \end{tabular}}
    \vspace{-2mm}
    \caption{Comparison with previous work on \textbf{Phoenix-2014} with WER as the evaluation metric.}
    \label{tab:sota_phoenix2014}
\end{table}

\subsection{Analysis and Ablation Study}
We first report the performance of our ISLR model. Next, we show the effectiveness of our proposed cross-lingual mapping by comparing it with a monolingual baseline and a multi-task training method. Then we compare the results of using different sign mappings mentioned in Section~\ref{sec:sign_map}. Besides, we study the influence of the number of mapped auxiliary signs and the sampling ratio. To save computational costs, we mainly conduct our ablation studies on CSL-Daily$\rightarrow$ Phoenix-2014T using SingleStream network with RGB inputs. Last, we show the advantage of our method in a data-scarcity scenario when using DGS as the auxiliary dataset and CSL as the primary dataset.

\begin{table}[t!]
\setlength\tabcolsep{2pt} 
    \centering
    \resizebox{\linewidth}{!}{ 
    \begin{tabular}{l c cccc c cccc}
    \toprule
           \multirow{2}{*}{Dataset} && \multicolumn{4}{c}{Dev} && \multicolumn{4}{c}{Test} \\
          \cmidrule{3-6} \cmidrule{8-11}
        && \multicolumn{2}{c}{Per-instance} & \multicolumn{2}{c}{Per-class} && \multicolumn{2}{c}{Per-instance} & \multicolumn{2}{c}{Per-class} \\
       &&top-1&top-5&top-1&top-5&&top-1&top-5&top-1&top-5\\
       \midrule
       Phoenix-2014T&&81.4&94.3&59.6&83.2&&81.7&94.3&60.9&82.4\\
       CSL-Daily&&73.6&84.5&73.0&90.4&&73.3&85.1&70.5&90.5\\
    \bottomrule
    \end{tabular}}
    \vspace{-2mm}
    \caption{Performance of our ISLR (isolated sign language recognition) model on Phoenix-2014T and CSL-Daily.}
    \label{tab:perf_islr}
\end{table}

\begin{table}[t!]
\setlength\tabcolsep{5pt} 
    \centering
    \resizebox{0.99\linewidth}{!}{ 
    \begin{tabular}{l c cc c cc}
    \toprule
     \multirow{2}{*}{Method}&& \multicolumn{2}{c}{SingleStream} && \multicolumn{2}{c}{TwoStream}  \\
     \cmidrule{3-4} \cmidrule{6-7} 
         &&Dev&Test&&Dev&Test\\
        \midrule
        Phoenix-2014T &&21.1&22.4&&17.7&19.3\\
        CSL-Daily, Phoenix-2014T&&20.7&21.9&&17.5&19.1\\
        \midrule
        CSL-Daily$\rightarrow$Phoenix-2014T&&\textbf{20.6}&\textbf{21.3}&&\textbf{17.3}&\textbf{18.6}\\ 
    \bottomrule
    \end{tabular}}
    \vspace{-2mm}
    \caption{Using two types of network architectures (SingleStream and TwoStream) proposed by~\cite{twostream_slr}, we compare our cross-lingual method with monolingual baseline and multi-task learning approach on Phoenix-2014T with WER as the evaluation metric.}
    \label{tab:ablation_cmp_baseline}
    \vspace{-5mm}
\end{table}

\vspace{-3mm}
\subsubsection{Isolated Sign Language Recognition}
\vspace{-1mm}
We employ a well-trained CSLR model to automatically partition the continuous signs into the isolated signs (Section~\ref{sec:dic_const}), which are then used to train an ISLR model (Section~\ref{sec:sign_map}). Although there is no ground truth for the segmentation, we can evaluate the ISLR model on the unseen dev and test samples as a proxy. 
We report the performance of our multilingual ISLR model on Phoenix-2014T and CSL-Daily in Table~\ref{tab:perf_islr} with per-instance/class top-1/5 accuracy as evaluation metrics. For both datasets, the per-instance top-1 accuracy exceeds $70\%$, implying that our automatic segmentation is able to spot video clips of the same sign word to build the dictionary, which provides a precondition for our subsequent cross-lingual CSLR training. The per-class accuracy for both datasets is much lower than the per-instance accuracy due to the imbalanced vocabulary distribution, particularly for Phoenix-2014T.
\vspace{-3mm}
\subsubsection{Effectiveness of Cross-lingual Sign Mapping} 
\vspace{-1mm}
To demonstrate the advantage of the proposed cross-lingual training methodology, we compare our CSL-Daily$\rightarrow$Phoenix-2014T model with two baselines:
\begin{itemize}
    \vspace{-2mm}
    \item \textit{Phoenix-2014T only.} We train a monolingual CSLR model with the identical network architecutre using only Phoenix-2014T.
    \vspace{-2mm}
    \item \textit{Multi-task learning on CSL-Daily and Phoenix-2014T.} We directly incorporate the original CSL-Daily dataset into the training of Phoenix-2014T without the sign mapping operation and append another CSL-Daily classification layer on top of the CSLR model, forming a multi-task learning framework. 
\end{itemize}
\vspace{-2mm}
We compare these two baselines with our method using both SingleStream and TwoStream models proposed by~\cite{twostream_slr}, with other configurations held the same. As shown in Table~\ref{tab:ablation_cmp_baseline}, multi-task learning (CSL-Daily, Phoenix-2014T) surpasses the monolingual baseline. Our multilingual method (CSL-Daily$\rightarrow$Phoenix-2014T) further reduces the WER for both architectures. This suggests that cross-lingual signs identified by our approach indeed enrich the training sources and hence facilitate the recognition capability of the primary dataset.

\begin{table}[t!]
\setlength\tabcolsep{5pt}
    \begin{subtable}[t]{0.2\textwidth}
        \centering
        \begin{tabular}{l c rr}
        \toprule
        Level && Dev & Test \\
        \midrule
        Instance && 20.8 & 21.5 \\
        Class && \textbf{20.6} & \textbf{21.3} \\
        \bottomrule
       \end{tabular}
       \caption{Instance level vs class level.}
       \label{tab:ablation_level}
    \end{subtable}
    \hspace{6mm}
    \begin{subtable}[t]{0.2\textwidth}
        \centering
        \begin{tabular}{l c rr}
        \toprule
        Mapping && Dev & Test \\
        \midrule
        DP && 20.7 & 21.9 \\
        CLP && \textbf{20.6} & \textbf{21.3} \\
        \bottomrule
       \end{tabular}
       \caption{DP: Dot-product, CLP: Cross-lingual prediction.}
       \label{tab:ablation_mapping}
    \end{subtable}

     \caption{Comparison between different cross-lingual mapping strategies by using SingleStream network under CSL-Daily$\rightarrow$ Phoenix-2014T setting. }
    \label{tab:ablation_matching_strat1}
    \vspace{-4mm}
\end{table}
\vspace{-2mm}
\subsubsection{Sign Mapping Strategies}
\label{sec:ablation_signmap}
\vspace{-1mm}
We describe different sign mapping strategies in Section~\ref{sec:sign_map}, including  1) \emph{class-level} versus \emph{instance-level}; 2) \emph{cross-lingual prediction} versus \emph{dot-product of weight matrices}. Now we experiment with these variants to show which strategy performs the best. We report their results in Table~\ref{tab:ablation_matching_strat1}. First, Table~\ref{tab:ablation_level} shows that when using cross-lingual prediction, the class-level mapping is superior to the instance-level mapping. This may be because averaging similarity scores over all instances belonging to the same sign could mitigate the noise raised by our imprecise segmentation of isolated sign videos. Second, Table~\ref{tab:ablation_mapping} shows that ``cross-lingual prediction'' outperforms ``dot-product of weight matrices'' for the class-level mapping. Therefore, we adopt the class-level mapping produced by the cross-lingual prediction as our default sign mapping strategy.
\vspace{-4mm}
\subsubsection{Similarity Threshold}
\begin{figure}[t!]
    \centering \vspace{-8mm}
    \includegraphics[width=1.0\linewidth,trim={0.3cm 0cm 0cm 0cm},clip]{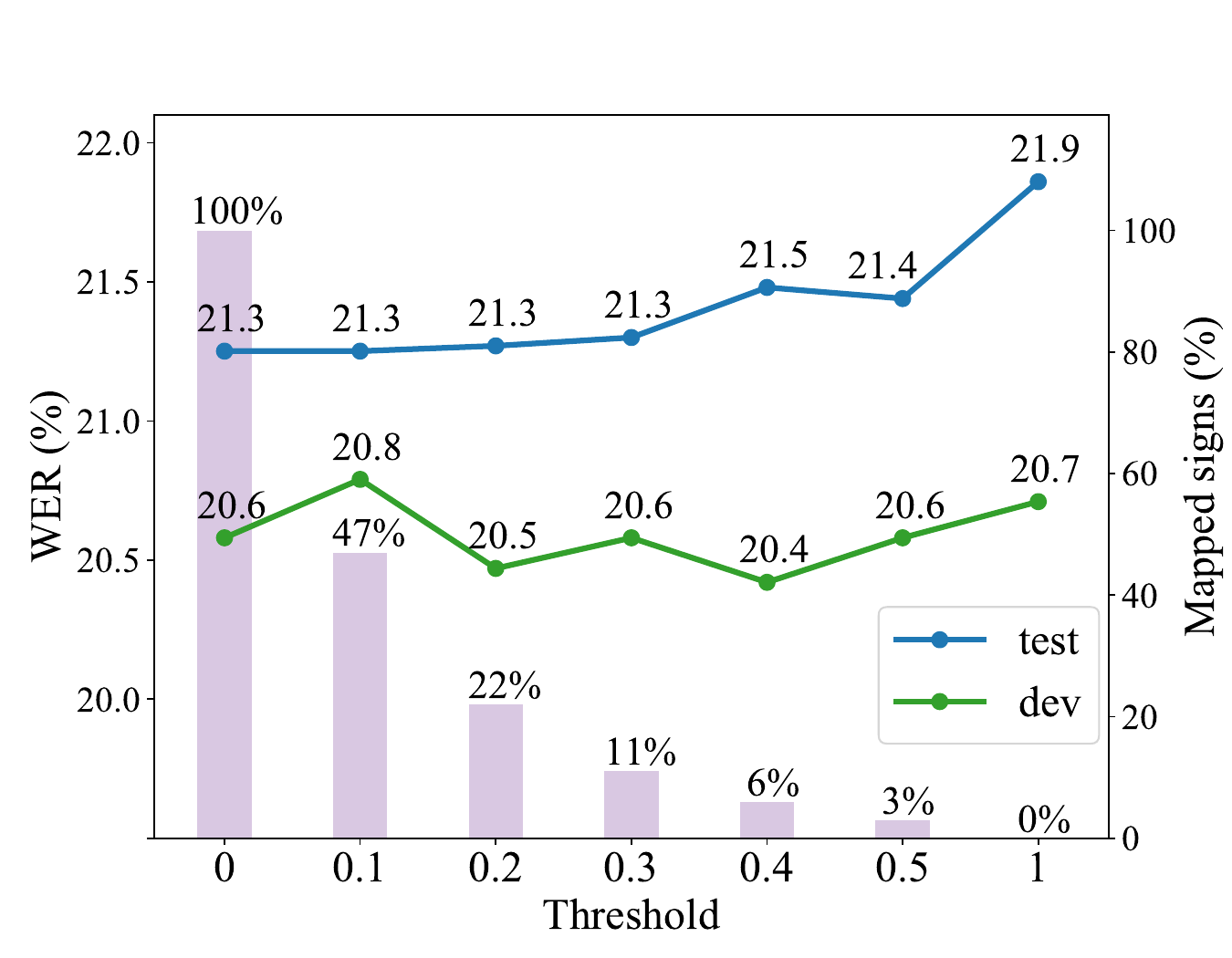}
    \vspace{-9mm}
    \caption{Effects of reducing the number of mapped signs by varying the threshold on the CSL-Daily$\rightarrow$ Phoenix-2014T setting. The bars show the ratio of mapped signs and the solid lines show the WER on Phoenix-2014T dev and test set (lower is better). }
    \label{fig:ablation_thresh}
\end{figure}

\vspace{-2mm}
As mentioned in Section~\ref{sec:sign_map}, for a sign mapping $\mathcal{M}_{A\rightarrow P}(s)$, we can use its maximum probability value, \textit{i.e.} the confidence of the cross-lingual prediction, as a proxy to estimate how similar the pair of cross-lingual signs are. The higher the confidence is, the more similar the pair of signs are. This suggests we may need to avoid mapping some auxiliary signs of low confidence. To study this, we set a confidence threshold for cross-lingual mapping. If the mapping confidence is higher than the threshold, we map the auxiliary sign to its cross-lingual sign. Otherwise, we preserve its original label for CSLR training.  We reduce the number of mapped signs by raising the threshold from 0 to 0.5 to see how the result changes accordingly, and plot the results in Figure~\ref{fig:ablation_thresh}. Note that a threshold of 1 preserves all auxiliary signs and becomes the multi-task learning baseline, \textit{i.e.} ``CSL-Daily, Phoenix-2014T'' in Table~\ref{tab:ablation_cmp_baseline}. We find that a majority of cross-lingual mappings have low confidence scores. However, our cross-lingual training method is insensitive to the change of the confidence threshold — any threshold between 0 and 0.3 outperforms the multi-task learning baseline on the test set. Hence, we map all the auxiliary signs and set the threshold to zero by default.

\begin{table}[t!]
\setlength\tabcolsep{6pt} 
    \centering
    \resizebox{\linewidth}{!}{ 
    \begin{tabular}{l c ccccccc}
    \toprule
     Ratio ($\alpha$) &&0&0.1&0.2&0.3&0.4&0.6&0.8 \\
    \midrule
     Dev  &&21.1&20.8&20.6&21.0&20.4&21.1&21.3 \\
     Test &&22.4&21.6&21.3&21.3&21.9&21.7&22.1 \\
    \bottomrule
    \end{tabular}}
    \vspace{-2mm}
    \caption{Ablation on different sampling ratios between the auxiliary dataset and the primary dataset. The experiment is conducted on CSL-Daily$\rightarrow$ Phoenix-2014T.}
    \label{tab:ablation_ratio}
    \vspace{-3mm}
\end{table}

\begin{table}[t!]
\vspace{2mm}
    \setlength{\tabcolsep}{5pt}
    \centering
    \small
    \begin{tabular}{@{}l  cc  cc  cc@{}}
    \toprule
      \multirow{2}{*}{Auxiliary}&\multicolumn{2}{c}{20\% CSL} & \multicolumn{2}{c}{40\% CSL} & \multicolumn{2}{c}{60\% CSL} \\
     & Dev&Test&Dev&Test&Dev&Test\\
    \midrule
    - & 45.2 & 45.5 & 36.1 & 35.9 & 32.7 & 32.8 \\
    Phoenix-2014T & 44.0 & 43.2 & 35.5 & 35.7 & 31.9 & 31.6 \\
    Phoenix-2014 & 44.0 & 44.5 & 35.3 & 35.2 & 31.9 & 32.0 \\
    \bottomrule
   \end{tabular}
   \vspace{-2.5mm}
   \caption{We compare the baseline without using auxiliary data, with Phoenix-2014/2014T$\rightarrow$20/40/60\% CSL-Daily.
   }
   \label{tab:csl_subset}
   \vspace{-4mm}
\end{table}

\vspace{-4mm}
\subsubsection{Ratio of Auxiliary Dataset} \vspace{-2mm}
\label{sec:ablation_ratio}
Here we study the effect of the sampling ratio between $\mathcal{D}_{A}$ and $\mathcal{D}_{P}$, \textit{i.e.} the coefficient $\alpha$ in Eq~\ref{eq:m_ctc}. We train our CSLR models with various $\alpha$ and compare their results in Table~\ref{tab:ablation_ratio}. Almost all models involving cross-lingual training surpass the monolingual baseline ($\alpha=0$).
\vspace{-3mm}

\subsubsection{Alleviating data-scarcity}
While CSL$\rightarrow$ DGS improves over training on DGS alone, we do not observe improvements in experiments of DGS$\rightarrow$ CSL. We suspect this is because the whole CSL dataset is larger in scale than the two DGS datasets and benefits little from auxiliary DGS samples. To prove the effectiveness of our method particularly when the primary dataset is scarce, we sample a subset of CSL-Daily as the primary dataset, use a DGS dataset as the auxiliary dataset and conduct the DGS$\rightarrow$ CSL experiments. As shown in Table~\ref{tab:csl_subset}, leveraging either Phoenix-2014T or Phoenix-2014 has an advantage over training on the CSL subset alone, bringing the largest gain when only 20\% of CSL is available. 

%% file: sections/conclusion.tex
\section{Conclusion}
\vspace{-1.5mm}
We present a novel approach to improve the monolingual performance of continuous sign language recognition (CSLR) by leveraging multilingual corpora and identifying visually similar signs across different sign languages, known as cross-lingual signs. Our method begins by constructing isolated sign dictionaries from the CSLR datasets. Next, we train a multilingual isolated sign recognition classifier on the two dictionaries and use it to identify the cross-lingual sign-to-sign mappings. Finally, we train our CSLR model on both the primary dataset and the remapped auxiliary dataset. By addressing the data scarcity issue, our approach achieves state-of-the-art results on two CSLR benchmarks. We are the first to demonstrate the effectiveness of cross-lingual transfer in CSLR and hope that our work will offer valuable insights for future research.

%% file: sections/appendix.tex

\section{Implementation Details}

\subsection{Dictionary Construction}
In Section 3.1, we employ a pre-trained CSLR model to partition the continuous sign videos into isolated sign clips. Here we describe the partition algorithm as follows.

\begin{figure*}[!t]
    \centering
        \includegraphics[trim={5cm 8cm 0cm 0cm},clip,width=0.9\textwidth]{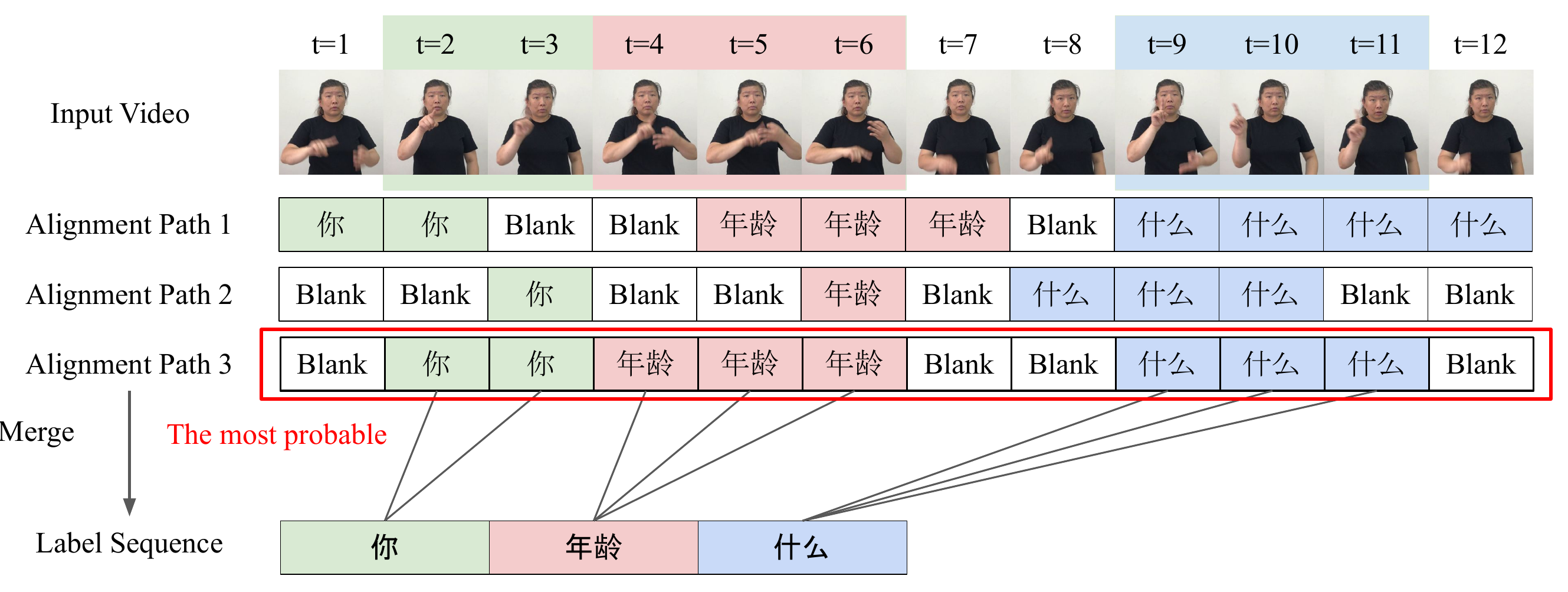}
    \caption{Illustration of partitioning a continuous sign video into the isolated sign clips. Given an input video $\boldsymbol{v}$ and its associated ground-truth sign sequence $\boldsymbol{s}$, we show three possible alignment paths (i.e. Alignment Path-1/2/3) with respect to $\boldsymbol{s}$. The probability of each alignment path can be computed by Eq.~\ref{eq:path_prob}. The optimal alignment path is the one with the maximal probability. After removing blank predictions and deduplicating the repeated non-blank predictions from the optimal alignment path, we could partition the input video into a collection of isolated sign clips.}
    \label{fig:merge}


\end{figure*}

To construct a dictionary $\mathcal{C}$ for a dataset $\mathcal{D}$ with an alphabet $\mathcal{S}$. Given a training video $\boldsymbol{v}=(\boldsymbol{v}_1,\ldots,\boldsymbol{v}_T) \in \mathcal{D}$ containing $T$ frames and its associated ground-truth sign sequence $\boldsymbol{s}=(s_1,\ldots, s_N), s_i\in \mathcal{S}$, a well-trained CSLR model produces a frame-wise prediction sequence $\boldsymbol{y}=(\boldsymbol{y}_1,\ldots,\boldsymbol{y}_T)$ in which $\boldsymbol{y}_t\in \mathbb{R}^{|\mathcal{S}'|}$ is a probability distribution over the expanded alphabet $\mathcal{S}'=\mathcal{S}\cup \{blank\}$ for the $t$-th frame\footnote{Since there is a downsampling layer in our CSLR network, the length of the output sequence is $T/4$. We temporarily upsample it by a factor of four to match the length of input $\boldsymbol{v}$.}. 
Therefore, the probability of a frame-wise sequence ${\boldsymbol{\pi}_{1:T}}=(\pi_1,\ldots,\pi_T)$ where $\pi_t \in \mathcal{S}'$, can be computed as
\begin{equation}
\label{eq:path_prob}
    p(\boldsymbol{\pi}_{1:T}\mid\boldsymbol{v})=\prod_{t=1}^{T}\boldsymbol{y}_t(\pi_t),
\end{equation}
where $\boldsymbol{y}_t(\pi_t)$ indicates the probability of observing label $\pi_t$ at timestamp $t$. 

A frame-wise sequence $\boldsymbol{\pi}_{1:T}$ can be mapped to a sign sequence by removing \textit{blank} predictions and deduplicating the repeated non-blank predictions. For a label sequence $\boldsymbol{s}$, we use $\Pi(\boldsymbol{s})$ to denote the set of frame-wise sequences that are mapped to $\boldsymbol{s}$ and call $\boldsymbol{\pi}_{1:T} \in \Pi(\boldsymbol{s})$ as an alignment path of $\boldsymbol{s}$. We illustrate the relationship between the label sequence $\boldsymbol{s}$ and its possible alignment paths $\boldsymbol{\pi}_{1:T}$ in Figure~\ref{fig:merge}. Now we need to find the optimal alignment path $\boldsymbol{\pi}_{1:T}^*$ as
\begin{equation}
    \label{eq:dtw_target}
\boldsymbol{\pi}_{1:T}^*=\argmax_{\boldsymbol{\pi}_{1:T}\in\Pi(\boldsymbol{s})}p(\boldsymbol{\pi}_{1:T}\mid\boldsymbol{v}).
\end{equation}

 \begin{figure}[t!]
    \centering
        \includegraphics[trim={0cm 7cm 5cm 0cm},clip,width=0.77\textwidth]{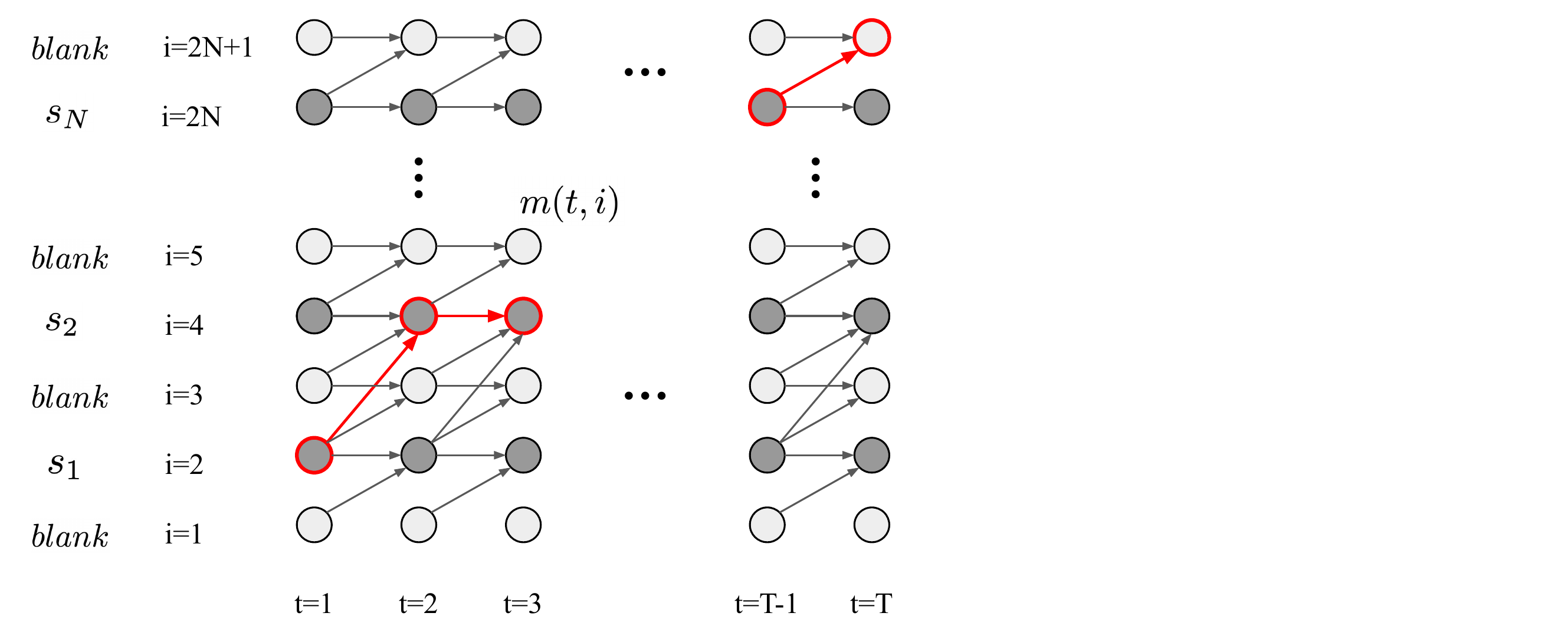}
    \caption{Illustration of the dynamic programming algorithm. Each node represents an intermediate variable $m(t,i)$  defined by Eq.~\ref{eq:m}. We iteratively compute the value of each node, as shown by the arrows. The probability of the optimal alignment path $\boldsymbol{\pi}_{1:T}^*$ is calculated by Eq.~\ref{eq:m_final}. After that, we could easily backtrack $\boldsymbol{\pi}_{1:T}^*$, as highlighted by the red nodes. Refer to Algorithm~\ref{alg:partition} for the whole process.}
    \label{fig:dtw}\vspace{-2mm} 

\end{figure}

\begin{algorithm}[t!]
\caption{Find the optimal alignment path}\label{alg:partition} 
\begin{algorithmic}
\State \textbf{Input:} frame-wise probabilities $\boldsymbol{y}$; extended label $\boldsymbol{s}'$ \State \textbf{Output:} the most probable alignment path $\boldsymbol{\pi}_{1:T}^*$
\For{$i\gets 1 \ \textrm{to} \ 2N+1 $} \Comment{Set the initial condition}
\If {$i\in\{1,2\}$}
    \State $m(1,i)=\boldsymbol{y}_1(s'_i)$
\Else 
    \State $m(1,i)=0$
\EndIf
\EndFor

\For{$i\gets 1 \ \textrm{to} \ 2N+1 $}
\Comment{Iterative computation}
        \If {$i=1$}
            \State $\mathcal{G}(i)=\{i\}$
        \ElsIf {$s'_i$ is $blank$ or $i=2$ or $s'_i=s'_{i-2}$}
            \State $\mathcal{G}(i)=\{i-1,i\}$
        \Else 
            \State $\mathcal{G}(i)=\{i-2,i-1,i\}$
        \EndIf
    \For{$t\gets 2 \ \textrm{to} \ T $} 
        \State $m(t,i)=\boldsymbol{y}_t(s'_i)\max_{j\in\mathcal{G}(i)}m(t-1,j)$        
    \EndFor
\EndFor
\State $i \gets \argmax_{j\in\{2N,2N+1\}}m(T,j)$ \Comment{Backtracking}
\State $\pi^*_{T} \gets  i$
\For{$t\gets T-1 \ \textrm{to} \ 1$}
\State $i\gets \argmax_{j\in\mathcal{G}(i)}m(t,i)$
\State $\pi^*_{t} \gets  i$
\EndFor
\State \textbf{return} $\boldsymbol{\pi}^*_{1:T}=(\pi^*_{1},\ldots,\pi^*_{T})$\
\end{algorithmic}

\end{algorithm}
\vspace{-2mm}

$\boldsymbol{\pi}_{1:T}^*$ can be efficiently searched by the dynamic time warping (DTW) algorithm~\cite{Müller2007DTW}. Formally, to accommodate \textit{blank} predictions in the alignment path, we first extend the label $\boldsymbol{s}$ of length $N$ to $\boldsymbol{s'}$ of length $2N+1$ by interleaving its items with $blank$:
\begin{equation*}
    \boldsymbol{s'}_{1:2N+1}=(blank,s_1,blank,s_2,\ldots,blank,s_N,blank).
\end{equation*}
In order to find the optimal path by Eq.~\ref{eq:dtw_target}, we define an intermediate variable $m(t,i)$ as the probability of the optimal path associated to the first $t$ frames of sign video $\boldsymbol{v}$ with sign sequence label $\boldsymbol{s'}_{1:i}$:

\begin{equation}
\label{eq:m}
m(t,i)=\max_{\boldsymbol{\pi}_{1:t}\in\Pi(\boldsymbol{s'}_{1:i})}p(\boldsymbol{\pi}_{1:t}|\boldsymbol{v}),
\end{equation}
where $p(\boldsymbol{\pi}_{1:t}|\boldsymbol{v})$ is formulated by Eq.~\ref{eq:path_prob}. Then the probability of the optical alignment path $\boldsymbol{\pi}_{1:T}^*$ can be calculated by:
\begin{equation}
\label{eq:m_final}
\max_{\boldsymbol{\pi}_{1:T}\in\Pi(\boldsymbol{s})}p(\boldsymbol{\pi}_{1:T}\mid\boldsymbol{v}) = \max_{j\in{\{2N,2N+1\}}}{m(T,j)}.
\end{equation}
Eq.~\ref{eq:m} can be computed recursively using dynamic programming (DP) as each $m(t, i)$ is a function of several earlier values. After obtaining the result of Eq.~\ref{eq:m_final}, we can seek out the optical alignment path $\boldsymbol{\pi}_{1:T}^*$ that gives rise to the maximum probability by backtracking. We illustrate the computation procedure in Figure~\ref{fig:dtw}. The details are also formulated in Algorithm~\ref{alg:partition}, which includes the initial condition, the Bellman equation for the DP algorithm, and how to backtrack the optical alignment path $\boldsymbol{\pi}_{1:T}^*$.

We find that among the estimated $\boldsymbol{\pi}_{1:T}^*$, many frames are predicted to be $blank$. For an isolated sign $s_i\in\boldsymbol{s}$ in the label sequence, if we only take the frames whose predictions in $\boldsymbol{\pi}_{1:T}^*$ are $s_i$ as the video clip for $s_i$, the resulting isolated video clips may be fairly short and not encompass the entire duration of that sign. To address this issue, we adopt the following strategy to find the video segment for 
$s_i\in \boldsymbol{s}$. First, we find the consecutive frames whose predictions are exactly $s_i$ in the optimal alignment path $\boldsymbol{\pi}_{1:T}^*$. Then, we expand their left and right boundaries by including more $blank$ frames whose predicted probability for $s_i$ is the highest when the \textit{blank} class is excluded. This approach yields an average length of 9 frames for an isolated segment. Table~\ref{tab:dataset_dict} shows the statistics of the constructed isolated sign dictionaries.

\begin{table}[t!]
\setlength\tabcolsep{5pt}
    \centering
    \resizebox{0.99\linewidth}{!}{
    \begin{tabular}{lp{0.1mm}ccccc}
    \toprule
        Dataset &\phantom{}&Frames&Train&Dev&Test&Vocab.\\
        \midrule
         Phoenix-2014~\cite{P2014}&&8.3&65,227&5,607&6,608&1231 \\
         Phoenix-2014T~\cite{camgoz2018neural}&&8.8&55,247&3,748&4,264&1085 \\
         CSL-Daily~\cite{zhou2021improving}&&9.0&133,714&8,173&9,002&2000 \\
    \bottomrule
    \end{tabular}}
    \caption{Statistics of the constructed isolated sign dictionaries produced by partitioning the continuous datasets (See Section 3.1). We show the average length of the segments (isolated signs), the number of segments in the Train/Dev/Test splits, and the vocabulary size for each dataset.}
    \label{tab:dataset_dict}

\vspace{-5mm}
\end{table}

\subsection{CSLR}
For continuous sign language recognition (CSLR), we re-use the architecture and training procedure of TwoStream-SLR~\cite{twostream_slr} except that we add an auxiliary dataset into the training dataset. We summarize our implementations as follows. 

\vspace{3mm}
\noindent\textbf{Architecture.} TwoStream-SLR~\cite{twostream_slr} contains two independent sub-networks to model RGB videos and estimated keypoint sequences. The keypoints are estimated by an HRNet~\cite{wang2020deep} trained on COCO-WholeBody~\cite{jin2020whole}. Each of the two sub-networks is an S3D~\cite{xie2018rethinking} backbone (only the first four blocks are used) pretrained on Kinetics-400~\cite{K400_dataset}. TwoStream-SLR also adopts bidirectional lateral connection, sign pyramid network and separate classification heads. Please refer to the original paper~\cite{twostream_slr} for more details.

\vspace{3mm}
\noindent\textbf{Training.} The training of our CSLR model consists of two stages. In the first stage, we separately pre-train the SingleStreamSLR-RGB/-keypoint using a single CTC loss~\cite{graves2006connectionist} without sign pyramid network and bidirectional lateral
connection. In the second stage, we load the pre-trained SingleStreamSLR networks and train the TwoStreamSLR using the CTC loss~\cite{graves2006connectionist} and a set of auxiliary losses proposed in~\cite{twostream_slr}. In each stage, we use the Adam optimizer~\cite{kingma_adam} with $\beta_1=0.9,\beta_2=0.998, \textrm{weight decay}=1e-3$ and a cosine learning scheduler to train the network for $40$ epochs with a batch size of 8 and a learning rate of $1e-3$. For our cross-lingual method, we mix $\mathcal{D}_{A\rightarrow P}$ and $\mathcal{D}_P$ with $\alpha=0.2$ defined in Equation 5. 

\vspace{3mm}
\noindent\textbf{Inference.} During inference, the final prediction is decoded into a sign sequence by CTC beam decoding~\cite{graves2006connectionist}. We use a beam width of 5. 

\subsection{ISLR}
Here we describe the architecture and training details of the isolated sign language recognition (ISLR) model we use for cross-lingual mapping.

\vspace{3mm}
\noindent\textbf{Architecture.} We adopt a TwoStream-ISLR architecture similar to the TwoStream-CSLR. The differences include: (1) the TwoStream-ISLR uses five blocks of the S3D network; (2) the sign pyramid networks are discarded; (3) a pooling layer is appended. 

\vspace{3mm}
\noindent\textbf{Training.}
The two S3D backbones in our TwoStream-ISLR are pre-trained on Kinetics-400~\cite{K400_dataset}. We train the whole network for $100$ epochs with a batch size of 32 and a learning rate of $1e-4$. We use the Adam optimizer~\cite{kingma_adam} with $\beta_1=0.9,\beta_2=0.998, \textrm{weight decay}=1e-3$ and a cosine learning schedule. We adopt the label smoothing with a smoothing weight of 0.2. We pad or truncate the input segments into the length of $16$ and apply augmentation including random spatial crop and random temporal sampling. We remove sign classes of frequency lower than $8$ for Phoenix-2014 and Phoenix-2014T and $20$ for CSL-Daily during training. This reduces their vocabulary size from 1231/1085/2000 to 428/389/981 respectively.

\vspace{3mm}
\noindent\textbf{Inference.}
During inference, we evenly pad or truncate input videos to the length of $16$. We forward samples of all classes to compute their cross-lingual predictions.

\begin{figure*}[h!]
     \centering
    \begin{subfigure}[t]{0.47\textwidth}
         \centering
        \includegraphics[trim={15cm 9cm 15cm 7cm},clip,width=\textwidth]{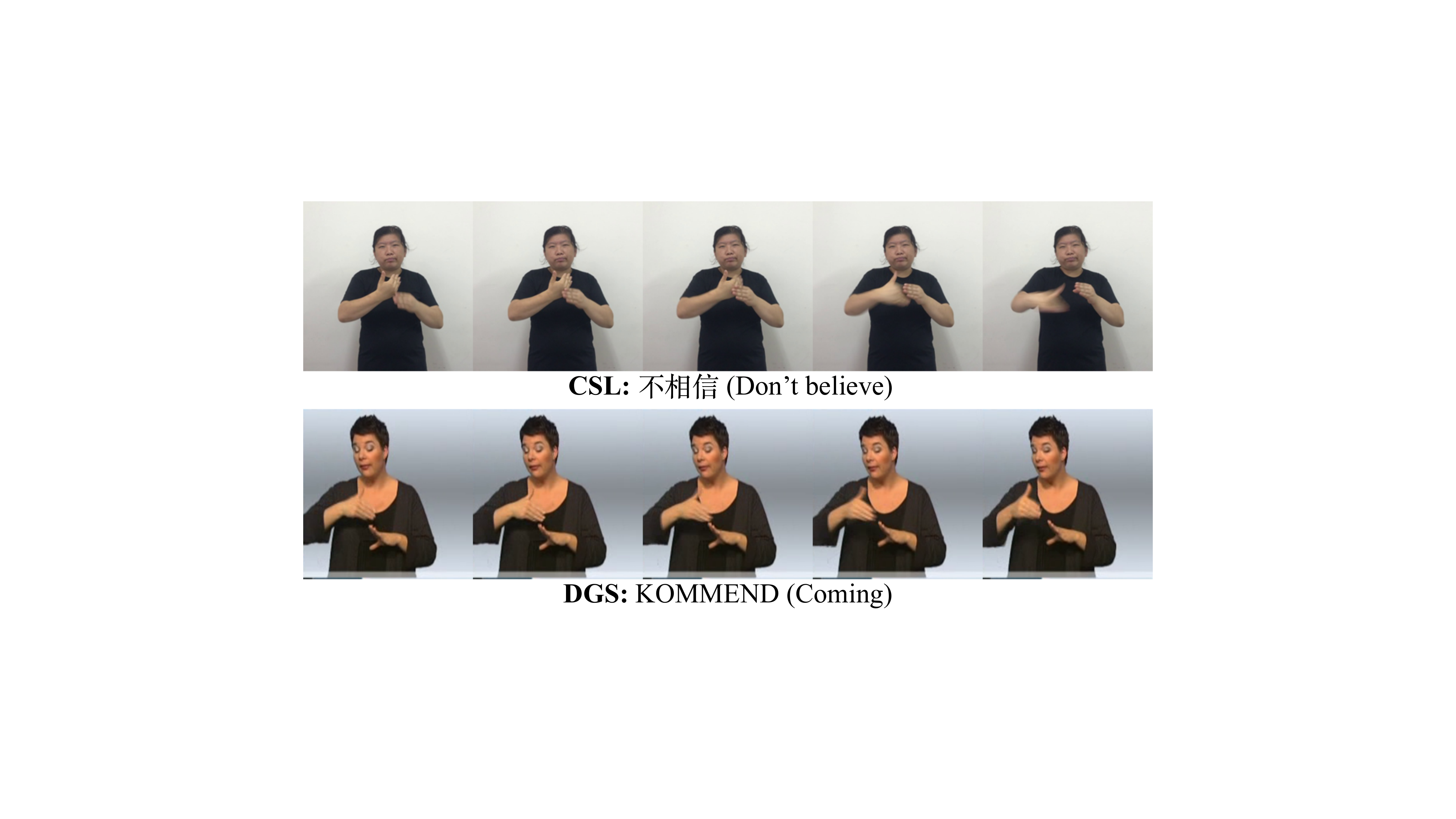}
        \vspace{-7mm}
        \caption{Confidence: 0.1}
         \label{fig:0.1}
     \end{subfigure}
    \hspace{8mm}
     \begin{subfigure}[t]{0.47\textwidth}
         \centering
        \includegraphics[trim={15cm 9cm 15cm 7cm},clip,width=\textwidth]{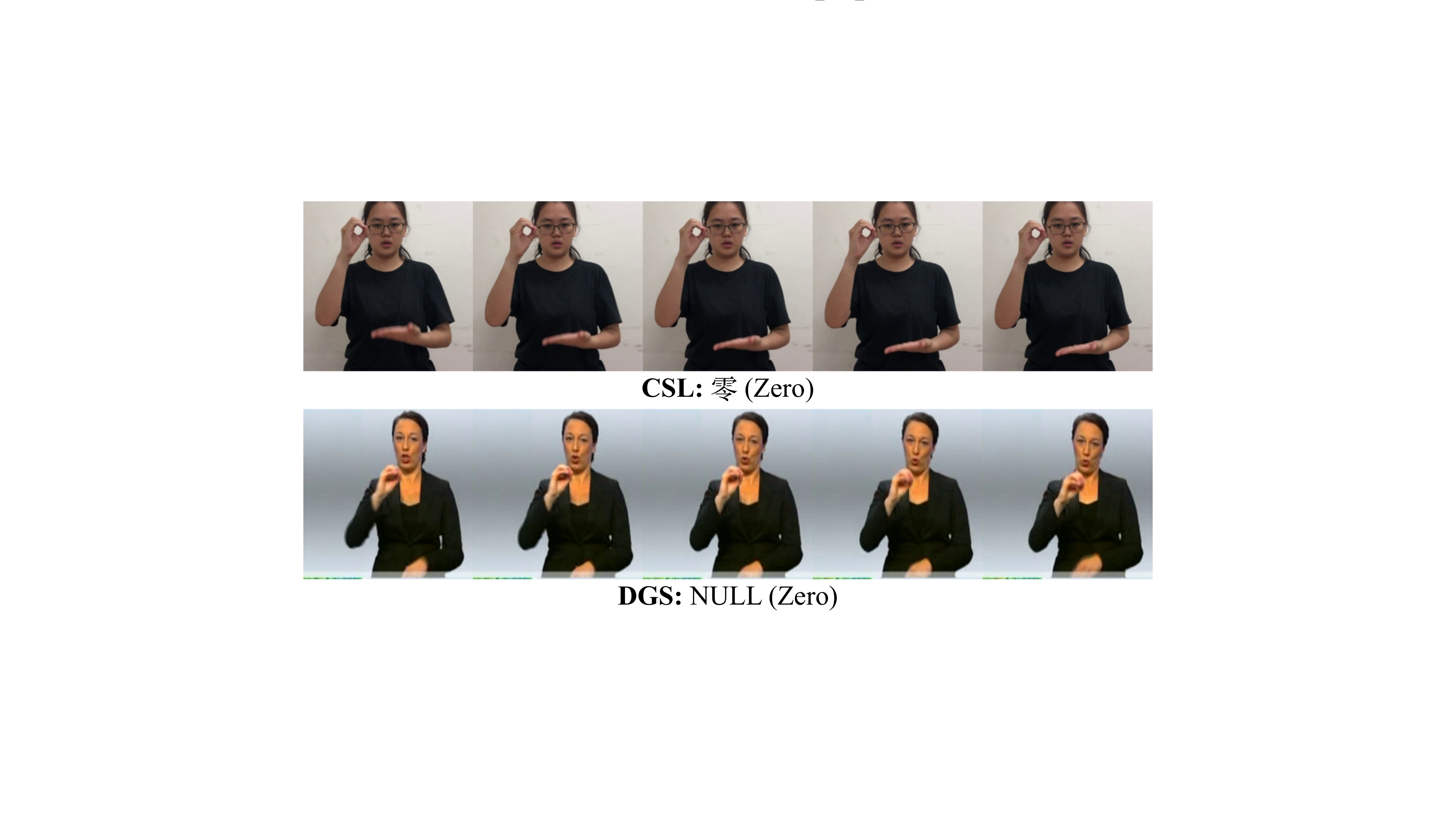}
        \vspace{-7mm}
        \caption{Confidence: 0.2}
         \label{fig:0.2}
     \end{subfigure}
    \begin{subfigure}[t]{0.47\textwidth}
         \centering
        \includegraphics[trim={15cm 9cm 15cm 7cm},clip,width=\textwidth]{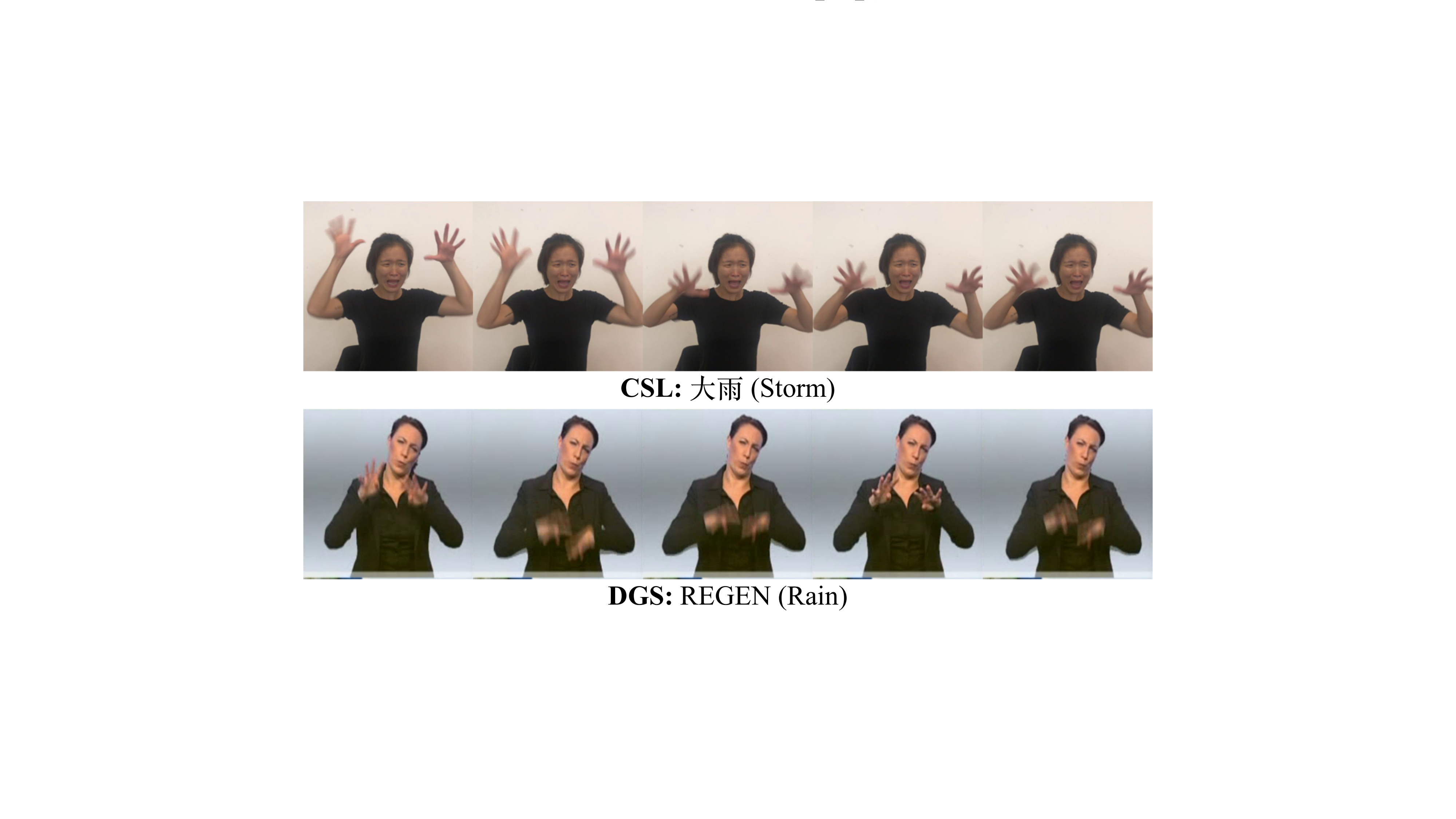}
        \vspace{-7mm}
        \caption{Confidence: 0.3}
         \label{fig:0.3}
     \end{subfigure}
    \hspace{8mm}
     \begin{subfigure}[t]{0.47\textwidth}
         \centering
        \includegraphics[trim={15cm 9cm 15cm 7cm},clip,width=\textwidth]{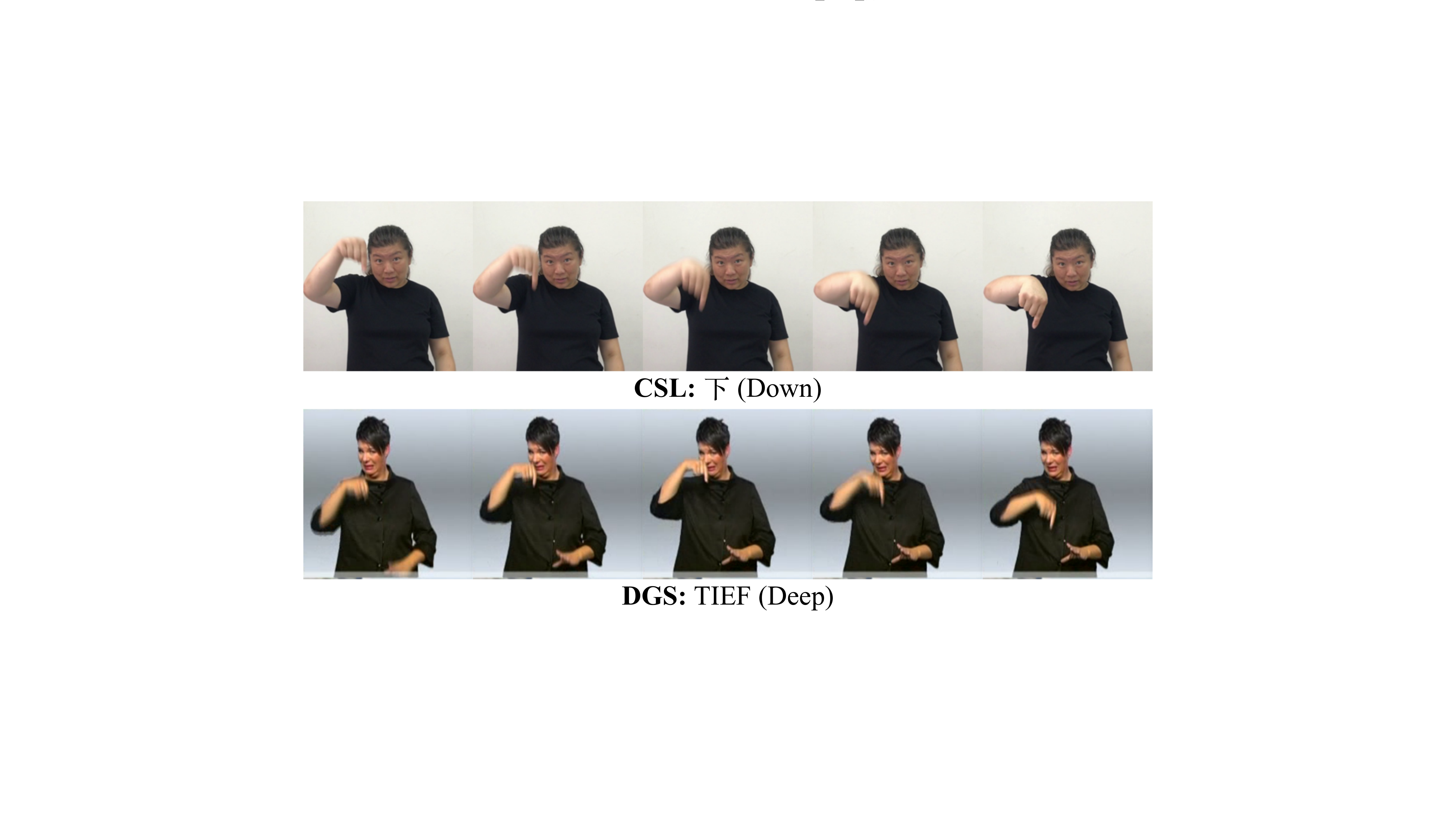}
        \vspace{-7mm}
        \caption{Confidence: 0.4}
         \label{fig:0.4}
     \end{subfigure}
    \hspace{8mm}
     \begin{subfigure}[t]{0.47\textwidth}
         \centering
        \includegraphics[trim={15cm 9cm 15cm 7cm},clip,width=\textwidth]{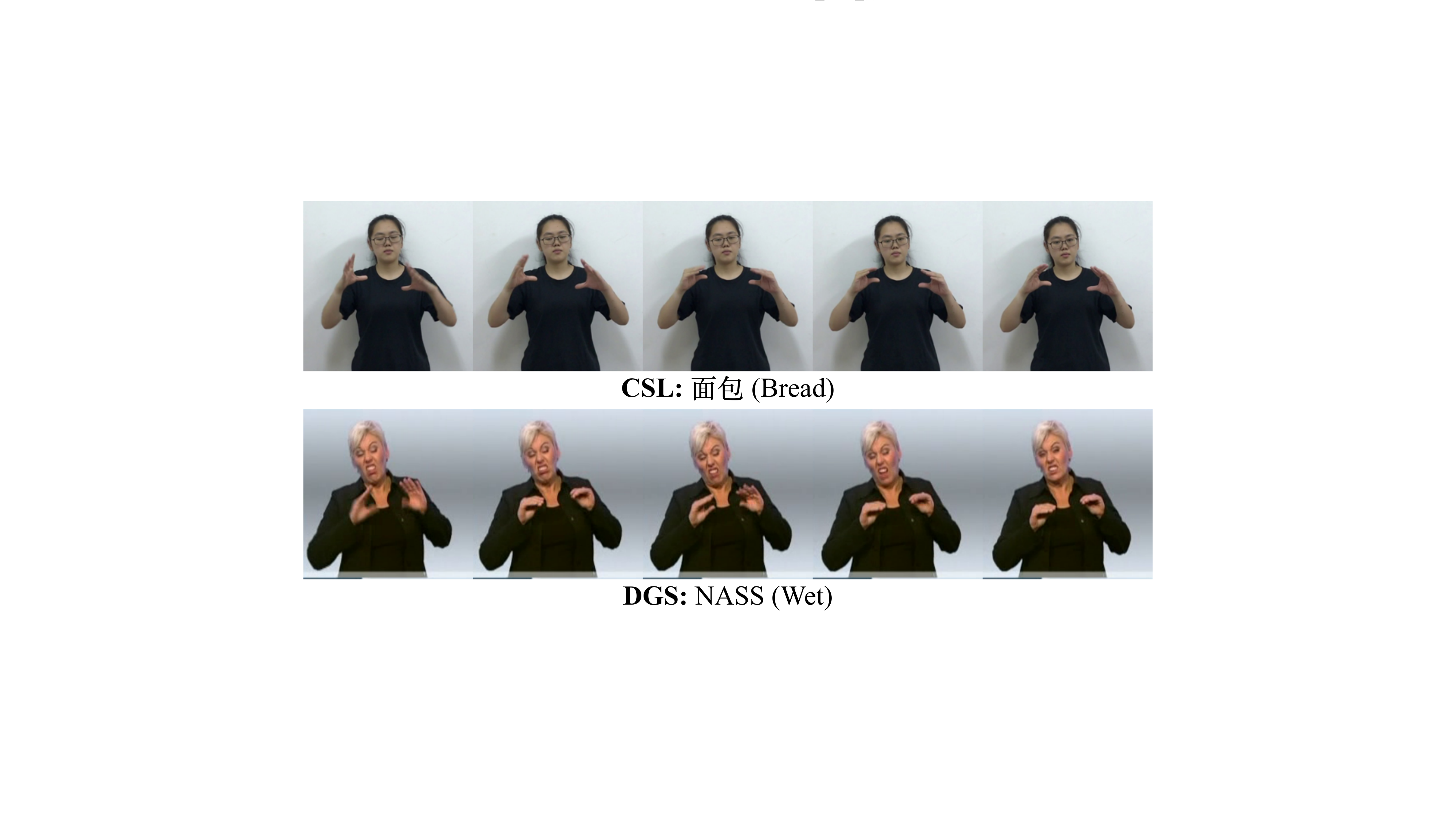}
        \vspace{-7mm}
        \caption{Confidence: 0.5}
         \label{fig:0.5}
     \end{subfigure}
    \hspace{8mm}
     \begin{subfigure}[t]{0.47\textwidth}
         \centering
        \includegraphics[trim={15cm 9cm 15cm 7cm},clip,width=\textwidth]{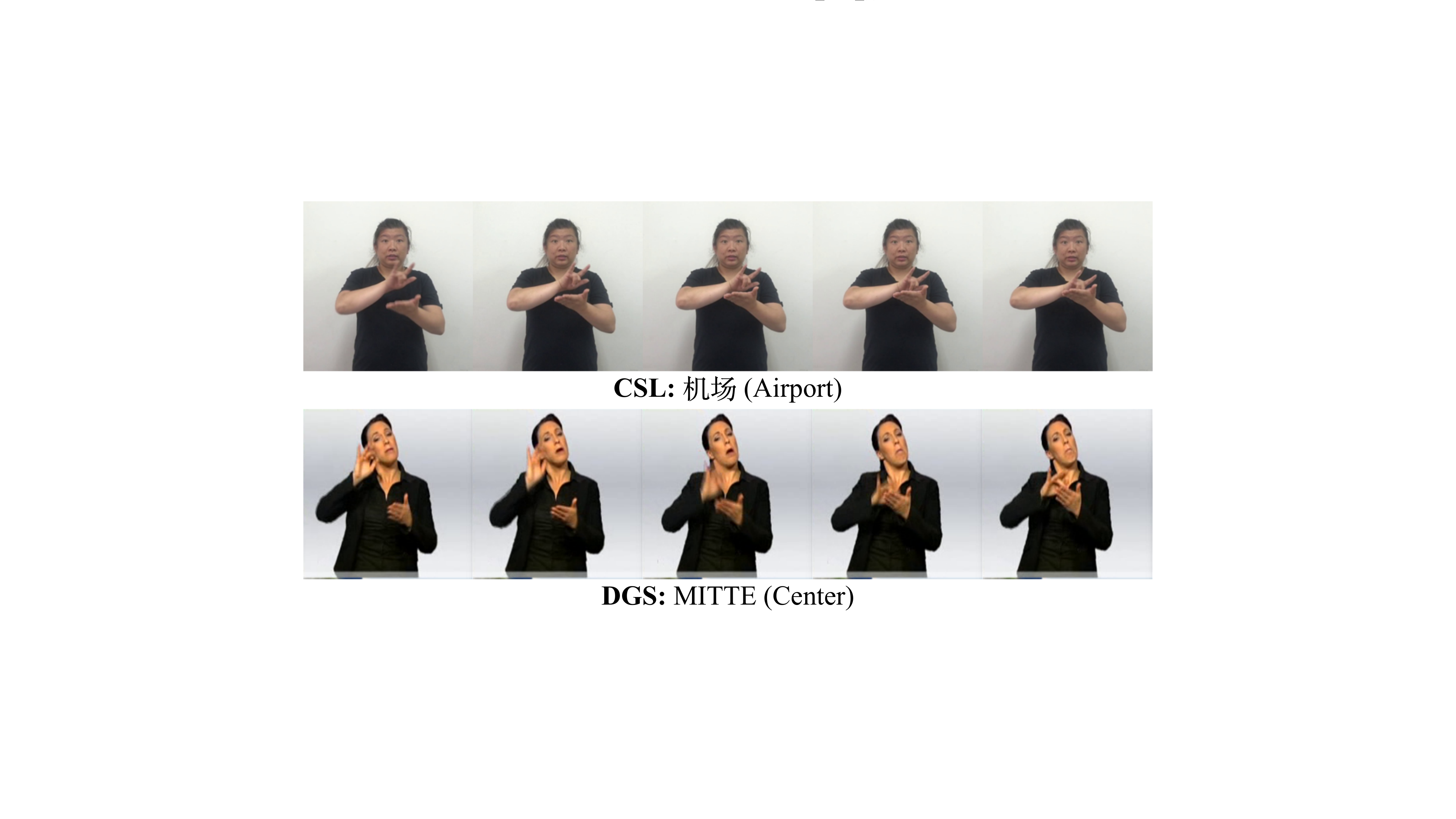}
        \vspace{-7mm}
        \caption{Confidence: 0.6}
         \label{fig:0.6}
     \end{subfigure}
    \hspace{8mm}
     \begin{subfigure}[t]{0.47\textwidth}
         \centering
        \includegraphics[trim={15cm 9cm 15cm 7cm},clip,width=\textwidth]{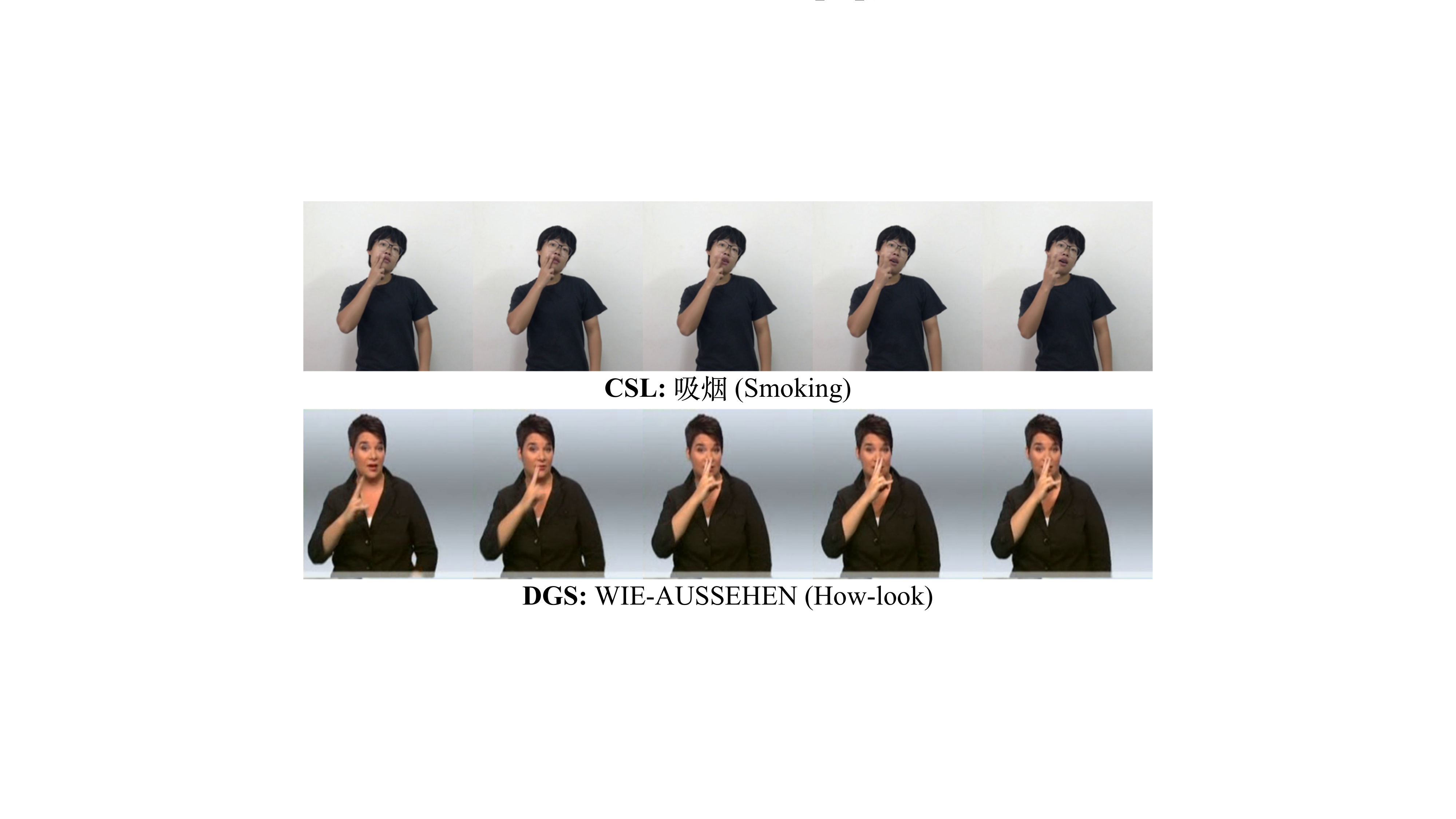}
        \vspace{-7mm}
        \caption{Confidence: 0.7}
         \label{fig:0.7}
     \end{subfigure}
    \hspace{8mm}
     \begin{subfigure}[t]{0.47\textwidth}
         \centering
        \includegraphics[trim={15cm 9cm 15cm 7cm},clip,width=\textwidth]{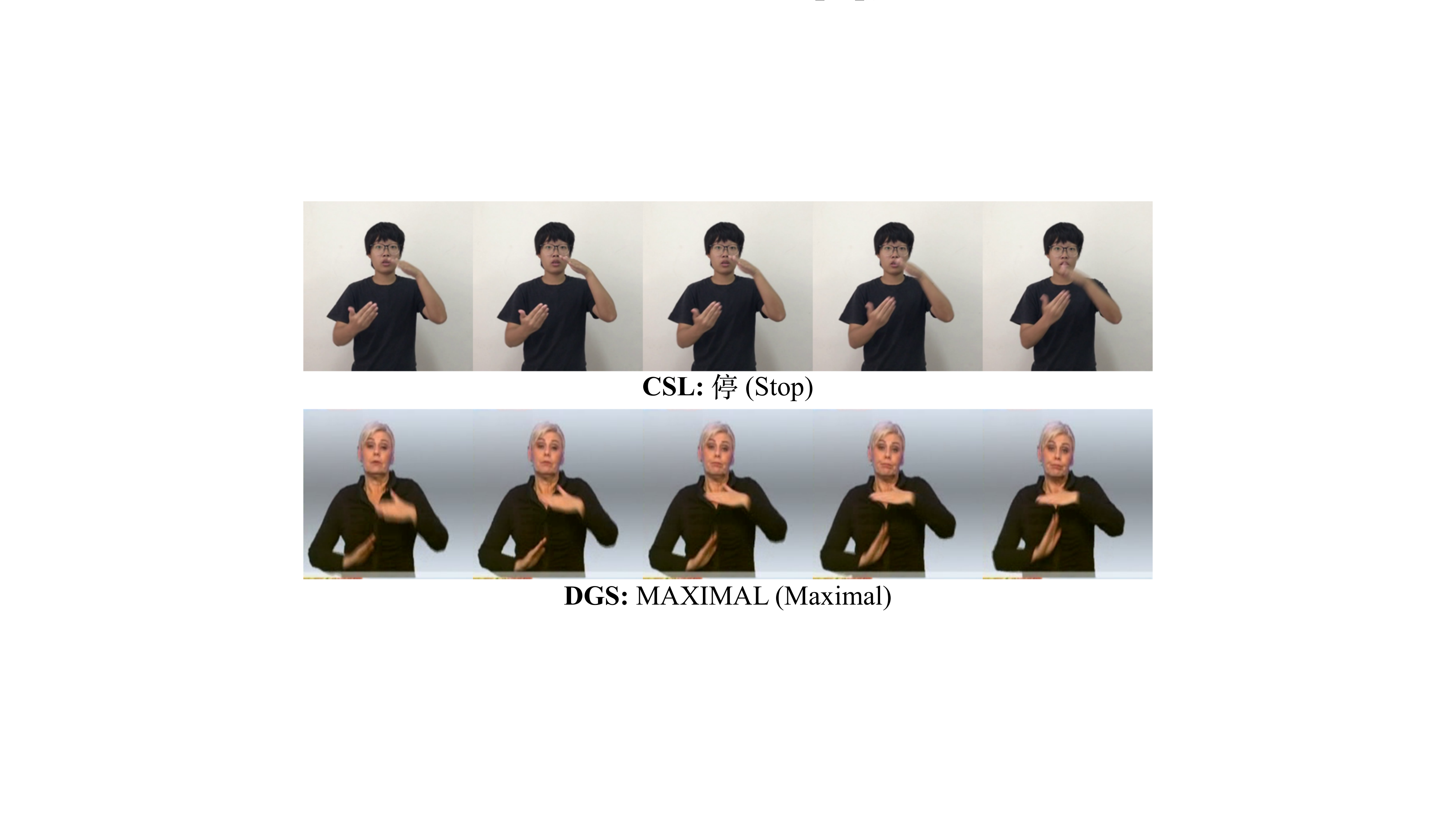}
        \vspace{-7mm}
        \caption{Confidence: 0.8}
         \label{fig:0.8}
     \end{subfigure}
    \caption{We show some examples of cross-lingual signs between Chinese sign language (CSL) and German sign language (DGS) using videos from CSL-Daily and Phoenix-2014T. We sort them by the cross-lingual prediction confidence. In general, higher confidence indicates higher similarity between the signs. Cross-lingual signs usually convey distinct meanings but occasionally share the same meaning, \textit{e.g.} 
    both express `zero' in Figure~\ref{fig:0.2}.}
    \vspace{-5mm}
    \label{fig:xlingual_sign_plus}
\end{figure*}

\section{Visualization of Cross-lingual Signs}
We illustrate more examples of the cross-lingual signs from CSL-Daily and Phoenix-2014T identified by our method in Figure~\ref{fig:xlingual_sign_plus}, where we sort the examples by their cross-lingual prediction confidences. 

First, we observe that all pairs of cross-lingual signs share similar visual cues, primarily the shape and movement of the hands. Furthermore, there appears to be a general trend where signs with higher confidence levels exhibit more detailed similarities. For example, in either Figure~\ref{fig:0.1} or Figure~\ref{fig:0.2}, the right hands of the two signers move similarly, while their left hands exhibit distinguishable patterns. In contrast, cross-lingual signs with confidence scores higher than 0.5, as depicted in Figure~\ref{fig:0.5}-\ref{fig:0.8}, not only share comparable hand orientations but also exhibit similar finger patterns and even facial expressions.

Next, cross-lingual signs usually carry distinct word meanings. For examples,  ``\begin{CJK*}{UTF8}{gbsn}面包\end{CJK*} (Bread)'' is mapped to ``KOMMEND(Coming)'' and ``\begin{CJK*}{UTF8}{gbsn}停\end{CJK*} (Stop)'' is mapped to ``MAXIMAL (Maximal)''. This demonstrates that DGS and CSL are mutually unintelligible. However, we also observe that some cross-lingual pairs convey identical meanings, \textit{e.g.} ``\begin{CJK*}{UTF8}{gbsn}零\end{CJK*} (Zero)'' and ``NULL(Zero)'', or close meanings, \textit{e.g.} ``\begin{CJK*}{UTF8}{gbsn}下\end{CJK*} (Down)'' and ``TIEF(Deep)''. This interestingly suggests that different deaf communities may share a common understanding of some semantic concepts regardless of their cultural and geographical difference and thus invent similar visual cues to convey some meanings.

\section{Discussion}
\noindent\textbf{Limitations and Future Directions.} Although our method is the first to demonstrate the effectiveness of cross-lingual transfer in CSLR, it requires both the primary dataset and the auxiliary dataset to have sequence-level annotations. Due to the limited number of labeled CSLR datasets, we are currently only able to apply our cross-lingual method to two sign languages, namely CSL and DGS. However, in the future, we aim to expand our approach to encompass a wider range of languages as more CSLR datasets become available. Additionally, we are excited to explore ways to utilize more cross-lingual data that lack labels so as to further enrich the training sources.

\vspace{3mm}
\noindent\textbf{Broader Impacts.} With the variation in sign languages across different regions, it has been a challenge to develop recognition systems that can cater to the needs of various deaf communities. However, our findings show that despite these variations, visually similar signs can be leveraged to improve the performance of such systems. This is particularly beneficial for under-represented deaf communities that have low-resource training data. Furthermore, our work has the potential to contribute to the broader field of sign linguistics. By identifying the commonalities and differences between different sign languages, we can enhance cross-cultural communication among deaf communities.